\let\latexdocument\document
\let\latexenddocument\enddocument

\documentclass[]{clv3custom}

\let\document\latexdocument
\let\enddocument\latexenddocument

\makeatletter
\AtBeginDocument{%
  \if@filesw
    \immediate\openout\@mainqry=\jobname.qry
  \fi
}
\AtEndDocument{%
   \ifx\@biography\@empty\else{\par\ifbrief\vskip10pt\fi\biofont\noindent\@biography\par}\fi
   \immediate\closeout\@mainqry
      \ifodd\c@page\clearpage\thispagestyle{empty}\null\clearpage\else\clearpage\fi
}
\makeatother

\NewCommandCopy{\cnumdef}{\numdef}
\NewCommandCopy{\endcnumdef}{\endnumdef}
\let\numdef\relax \let\endnumdef\relax

\usepackage{listings}
\usepackage{array}
\usepackage{booktabs}
\usepackage{multirow}
\usepackage{multicol}
\usepackage{amsmath,amssymb,amsfonts}
\usepackage{diagbox}
\usepackage{enumitem}

\usepackage{xcolor}
\usepackage{keystroke}
\usepackage{hyperref}
\definecolor{darkblue}{rgb}{0, 0, 0.5}
\hypersetup{colorlinks=true,citecolor=darkblue, linkcolor=darkblue, urlcolor=darkblue}

\definecolor{valbest}{HTML}{d9ead3}
\newcommand{\valbest}[1]{\colorbox{valbest}{#1}}
\definecolor{valgood}{HTML}{d0e0e3}
\newcommand{\valgood}[1]{\colorbox{valgood}{#1}}
\definecolor{valmid}{HTML}{fce5cd}
\newcommand{\valmid}[1]{\colorbox{valmid}{#1}}
\definecolor{valbad}{HTML}{ead1dc}
\newcommand{\valbad}[1]{\colorbox{valbad}{#1}}
\definecolor{themegreen}{HTML}{365956}
\definecolor{themepurple}{HTML}{3c1b48}
\definecolor{themered}{HTML}{b43748}

\newcommand{\llm}[1]{LLM#1}
\newcommand{\rulesep}{\unskip\ \vrule\ }

\newcommand{\numClfTasks}{20}
\newcommand{\numGenTasks}{5}
\newcommand{\numTasks}{25}

\newcommand{\revision}[1]{#1}

\begin{document}
\issue{50}{1}{2024}

\dochead{}

\runningtitle{Can LLMs Transform CSS?}

\runningauthor{Ziems, Held, Shaikh, Chen, Zhang, Yang}

\pageonefooter{Action editor: Vivek Srikumar. Submission received: 26 April 2023; revised version received: 29 August 2023; accepted for publication: 25 October 2023.}

\title{Can Large Language Models Transform Computational Social Science?}

\author{Caleb Ziems\thanks{E-mail: Caleb Ziems: \texttt{cziems@stanford.edu}. William Held: \texttt{wheld3@gatech.edu}; Omar Shaikh: \texttt{oshaikh@stanford.edu}; Jiaao Chen: \texttt{jiaaochen@gatech.edu}; Zhehao Zhang: \texttt{hehao.zhang.gr@dartmouth.edu}; Diyi Yang: \texttt{diyiy@stanford.edu}}}
\affil{Stanford University}

\author{William Held}
\affil{Georgia Institute of Technology}

\author{Omar Shaikh}
\affil{Stanford University}

\author{Jiaao Chen}
\affil{Georgia Institute of Technology}

\author{Zhehao Zhang}
\affil{Dartmouth College}

\author{Diyi Yang\thanks{Contribution distributed as follows. Caleb, Will, and Diyi decided the project scope and research questions. Caleb performed the CSS literature review, as well as the subject and task selection. Will built the evaluation pipeline and prompting guidelines. Will, Caleb, and Omar all contributed data pre-processing, loading, and evaluation scripts. Caleb ran the OpenAI and Flan-T5/UL2 prompt perturbation experiments and few-shot experiments. Zhehao was responsible for all baseline experiments. Caleb managed the human evaluations. All authors contributed to discussion, results, error analysis, and paper writing.}}
\affil{Stanford University}

\maketitle

\begin{abstract}
Large Language Models (\llm{s}) are capable of successfully performing many language processing tasks zero-shot (without training data). If zero-shot \llm{s} can also reliably classify and explain social phenomena like persuasiveness and political ideology, then \llm{s} could \revision{augment the} Computational Social Science (CSS) pipeline in important ways. This work provides a road map for using \llm{s} as CSS tools. Towards this end, we contribute a set of prompting best practices and an extensive evaluation pipeline to measure the zero-shot performance of 13 language models on \numTasks{} representative English CSS benchmarks. On taxonomic labeling tasks (classification), \llm{s} fail to outperform the best fine-tuned models but still achieve fair levels of agreement with humans. On free-form coding tasks (generation), \llm{s} produce explanations that often \textit{exceed} the quality of crowdworkers' gold references. We conclude that the performance of today's \llm{s} can augment the CSS research pipeline in two ways: (1) serving as zero-shot data annotators on human annotation teams, and (2) bootstrapping challenging creative generation tasks \revision{(e.g., explaining the underlying attributes of a text)}. In summary, \llm{s} are posed to meaningfully participate in social science analysis in partnership with humans.
\end{abstract}

\section{Introduction}
\begin{quote}
    \textit{The most surprising scientific changes tend to arrive, not from accumulated facts and discoveries, but from the invention of new tools and methodologies that trigger ``paradigm shifts''} \citep{kuhn1962structure}. 
\end{quote}

\emph{Computational Social Science} (CSS) \citep{lazer2020computational} was born from the immense growth of human data traces on the web and the rapid acceleration of computational resources for processing this data. These developments allowed researchers to study language and behavior at an unprecedented scale \citep{lazer2009computational}, with both global and fine-grained observations \citep{golder2014digital}. From the early days of content dictionaries \citep{stone1966general}, statistical text analysis facilitated CSS research by providing structure to non-numeric data. Now, Large Language Models (\llm{s}) may be poised to change the computational social science landscape by providing such capabilities without custom training data.

\begin{figure*}[t!]
    \centering
    \includegraphics[width=\linewidth]{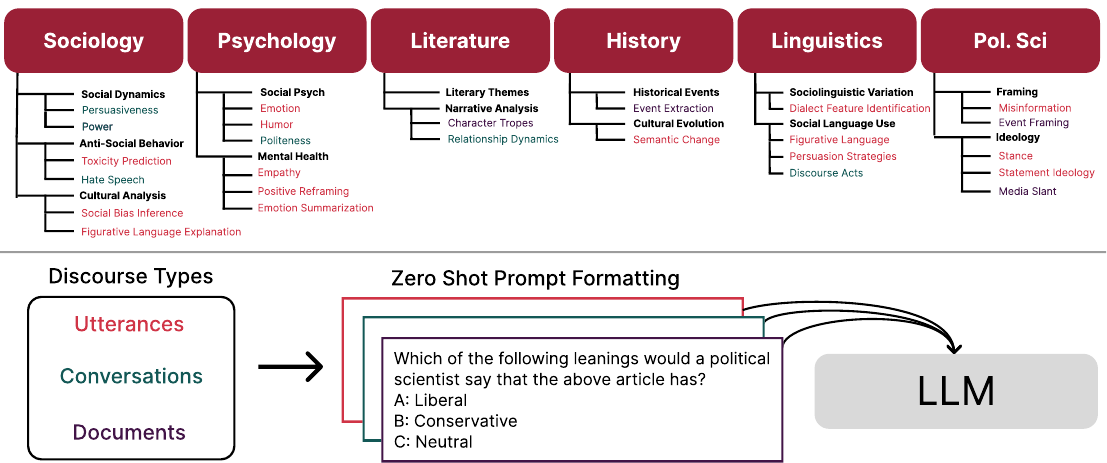}
    \caption{We assess the potential of \llm{s} as multi-purpose tools for CSS. We identify core subject areas in prior CSS work and select 24 diverse and representative tasks from across these fields (top). Then, we segment tasks into distinct discourse types and evaluate both open and closed-source \llm{s} across this benchmark using zero-shot prompting (bottom).}
    \label{fig:crown_jewel}
\end{figure*}

The goal of this work is to assess the degree to which \textit{\llm{s} can transform Computational Social Science (CSS).} Solid computational approaches are needed to help analyze textual data and to understand a variety of social phenomena across academic disciplines. Current CSS methodologies typically use \textit{supervised} text classification and generation in order to scale up manual \revision{labeling} efforts to unseen texts \revision{(also called \textit{coding} in the social sciences)}. Reliable supervised methods typically demand an extensive amount of human-annotated training data. Alternatively, \textit{unsupervised} methods can run ``for free,'' but the resulting output can be uninterpretable. In the status quo, data resources constrain the theories and subjects CSS can be applied to.

\llm{s} have the potential to remove these constraints. Recent \llm{s} have demonstrated the striking ability to reliably classify text, summarize documents, answer questions, and generate interpretable explanations in a variety of domains, even exceeding human performance \textit{without the need for supervision} \citep{bang2023multitask,qin2023chatgpt, zhuo2023exploring,goyal2022news}. If \llm{s} can similarly provide reliable labels and summary codes through zero-shot prompting, CSS research can be broadened to a wider range of hypotheses than current tools and data resources support. Zero-shot viability in this space is our primary research question. To effectively harness the power of \llm{s}, behavioral researchers should understand the pros and cons of different modeling decisions (model-selection), as well as how these decisions intersect with their fields of specialization
(domain-utility) and downstream use-cases (functionality). By evaluating \llm{s} on an extensive suite of CSS tasks, this work provides researchers a roadmap with answers to the following research questions:

\begin{itemize}
    \item \textbf{(RQ1) Viability:} Are \llm{s} able to augment the human annotation pipeline? Can they \revision{match or exceed the reliability of human annotation?}
    \item \textbf{(RQ2) Model-Selection:} How do different aspects of \llm{s} (e.g., model size, pretraining) affect their performances on CSS tasks? 
    \item \textbf{(RQ3) Domain-Utility:} Are zero-shot \llm{s} specially adapted for better results in some fields of science rather than others? 
    \item \textbf{(RQ4) Functionality:} Are zero-shot \llm{s} equipped to assist with labeling tasks (classification) or summary-explanatory tasks (generation) or both?
\end{itemize}

The research pipeline in Figure~\ref{fig:crown_jewel} allows us to answer these questions. First, we survey the social science literature to understand where \llm{s} could serve as analytical tools (\S\ref{sec:core_methods_css}). Then we operationalize each use-case with a set of representative tasks (\S\ref{sec:tasks}). Specifically, classification and parsing methods can help researchers code for linguistic, psychological, and cultural categories (\S\ref{subsec:utterances} - \S\ref{subsec:document})
while generative models can explain underlying constructs (e.g.,~figurative language, emotional reactions, hate speech, and misinformation),
and restructure text according to established theories like cognitive behavioral therapy (\S\ref{subsec:generation}). With a final evaluation suite of 24 tasks, we test the zero-shot performance of 13 language models with differing architectures, sizes, pre-training, and fine-tuning paradigms (\S\ref{subsec:clf_results}, \S\ref{subsec:gen_results}). This allows us to suggest actionable steps for social scientists interested in co-opting \llm{s} for research (\S\ref{sec:discussion}). Specifically, we suggest a blended supervised-unsupervised scheme for human-AI partnered labeling and content analysis.

Concretely, our analysis reveals that, except in minority cases, prompted \llm{s} do not match or exceed the performance of carefully fine-tuned classifiers, and the best \llm{} performances are often too low to entirely replace human annotation. However, \llm{s} \textit{can} achieve fair levels of agreement with humans on labeling tasks \revision{(RQ1)}. These results are not limited to a subset of academic fields, but rather span the social sciences across a range of conversation, utterance, and document-level classification tasks \revision{(RQ2)}. Furthermore, the benefits of \llm{s} are compounded as models scale up \revision{(RQ3)}. This suggests that \llm{s} can augment the annotation process through iterative joint-labeling, significantly speeding up and improving text analysis in the social sciences.

Importantly, some \llm{s} can also generate informative explanations for social science constructs. Leading models can achieve parity with the quality of dataset references, and can even exceed them in terms of relevance, coherence, faithfulness, and fluency. Humans prefer model outputs 50\% of the time, suggesting that human-AI collaboration will extend beyond labeling tasks to the joint coding of new constructs, analyses, and summaries.

\section{An Overview of CSS}
\label{sec:core_methods_css}
Following \citet{lazer2020computational}, we define Computational Social Science as the development and application of computational methods to the scientific analysis of behavioral and linguistic data. Critically, CSS centers around the scientific method, forming and testing broad and objective hypotheses, while similar efforts in the Digital Humanities (DH) focus more on the subjectivity and particularity of events, dialogues, cultures, laws, value-systems, and human activities \citep{dobson2019critical}. 

This section surveys the current needs of researchers in both the computational social sciences and digital humanities. We choose to merge our discussion under the banner of CSS, since solid computational approaches are needed to help analyze textual data and to understand a variety of socio-behavioral phenomena across both scientific and humanistic disciplines.
We focus primarily on the most tractable text classification, structured parsing, summarization, and natural language generation tasks for CSS. 
Some other techniques like aggregate mining of massive datasets or topic modeling may be largely outside the scope of transformer-based language models, which have a fixed processing window size and quadratic space complexity. 

The following subsections outline how computational methods can support specific fields of inquiry regarding how people think (\textit{psychology}; \S\ref{subsec:psychology}), communicate (\textit{linguistics}; \S\ref{subsec:linguistics}), establish governance and value-systems (\textit{political science, economics}; \S\ref{subsec:polisci}), collectively operate (\textit{sociology}; \S\ref{subsec:sociology}), and create culture (\textit{literature, anthropology}; \S\ref{subsec:literature}) across time (\textit{history}; \S\ref{subsec:history}).
\subsection{History} 
\label{subsec:history}
Historians study \textit{events}, or transitions between states \citep{box2004event,abbott1990conceptions}, like the onset of a war. Event extraction is a parsing task from unstructured text to more regular data structures which capture the location, time, cause, and participants in the event \citep{xiang2019survey}. This task, which is central to a growing number of computational studies on history \citep{lai-etal-2021-event, sprugnoli2019novel}, can be broken into (1) event detection, and (2) event argument extraction, which we benchmark in \S\ref{subsub:event_detection} and \S\ref{subsub:event_argument_extraction} respectively. Historians also work to understand the influence of events on historical shifts in \textit{discourse} \citep{dimaggio2013exploiting} and \textit{meaning} \citep{hamilton2016cultural}. We further discuss NLP for discourse and semantic change in \S\ref{subsec:polisci} and \S\ref{subsec:linguistics}.
\subsection{Literature}
\label{subsec:literature}
Literary studies are closely tied to the analysis of \textit{themes} \citep{jockers2013significant}, \textit{settings} \citep{piper2021narrative}, and \textit{narratives} \citep{sap2022quantifying,saldias-roy-2020-exploring,boyd2020narrative}. Settings can be identified using named entity recognition \citep{brooke-etal-2016-bootstrapped} and toponym resolution \citep{delozier-etal-2016-creating}, which are already demonstrably solved by prompted models like GPT 3.5 Turbo \citep{qin2023chatgpt}. Themes are typically the subject of topic modeling, which is outside the scope of \llm{s}.
Instead we focus on NLP for narrative analysis. NLP systems can be used to parse narratives into chains \citep{chambers-jurafsky-2008-unsupervised} with \textit{agents} \citep{coll-ardanuy-etal-2020-living,vala-etal-2015-mr} their \textit{relationships} \citep{labatut2019extraction,iyyer2016feuding,srivastava2016inferring}, and the \textit{events} \citep{sims2019literary} they participate in. We cover social role labeling and event extraction methods in \S\ref{subsub:role_tagging} and \S\ref{subsub:event_argument_extraction} respectively. Researchers can also study agents in terms of their \textit{power} dynamics \citep{sap2017connotation} and \textit{emotions} \citep{brahman-chaturvedi-2020-modeling}, which we benchmark in \S\ref{subsub:power} and \S\ref{subsub:emotion}. \textit{Figurative language} \citep{kesarwani2017metaphor} and \textit{humor} classification \citep{davies2017sociolinguistic} are two other relevant tasks for the study of literary devices, and we evaluate these tasks in \S\ref{subsub:figurative_language} and \S\ref{subsub:humor}.
\subsection{Linguistics} 
\label{subsec:linguistics}
Computational sociolinguists use computational tools to measure the interactions between society and language, including the stylistic and structural features that distinguish speakers \citep{nguyen2016computational}. Language variation is closely related to social identity \citep{bucholtz2005identity}, from group membership \citep{del2017semantic}, geographical region \citep{purschke2019lorres}, and social class \citep{preotiuc2015analysis,del2017semantic} to personal attributes like age and gender \citep{bamman2014gender}. In \S\ref{subsub:dialect_feature_detection} and \S\ref{subsub:semantic_change}, we use \llm{s} to identify the structural features of English dialects, which linguists can use to classify and systematically study dialects, measure different feature densities in different population strata, and study the onset and diffusion of language change \citep{kershaw2016towards,eisenstein2014diffusion,ryskina2020new,kulkarni2015statistically,hamilton2016diachronic,di2019training,zhu2021structure,schlechtweg-etal-2020-semeval}.
\subsection{Political Science} 
\label{subsec:polisci}
Political scientists study how political actors move \textit{agendas} \citep{grimmer2010bayesian} by persuasively \textit{framing} their discourse ``to promote a particular problem definition, causal interpretation, moral evaluation, and/or treatment recommendation'' \citep{entman1993framing}. These agendas cohere within \textit{ideologies}. Computational social scientists have advanced political science through the detection of political leaning, ideology, belief, and stance \citep{ahmed2010staying,baly2020we,bamman2015open,iyyer2014political,johnson2017ideological,preoctiuc2017beyond,luo2020desmog,stefanov2020predicting}, as well as \textit{issue} \citep{iyengar1990framing} and \textit{entity} framing \citep{van2020doctor}. Applications for persuasion, framing, ideology, and stance detection in the social sciences are numerous. Analysts can uncover fringe issue topics \citep{bail2014terrified} and
frames \citep{ziems2021protect,mendelsohn2021modeling,demszky2019analyzing,field2018framing}, with applications to public opinion \citep{bhatia2017associative,garg2018word,kozlowski2019geometry,abul2018solidarity}, voting behavior \citep{black2011emotions}, policy change \citep{flores2017anti}, social movements \citep{nelson2021cycles,sech2020civil,rogers2019calls,tufekci2012social}, and international relations \citep{king2003automated}. We benchmark ideology detection in \S\ref{subsub:utterance_ideology} and \S\ref{subsub:media_ideology}, stance detection in \S\ref{subsub:utterance_stance}, and entity framing in \S\ref{subsub:role_tagging}. Furthermore, understanding the discourse structure and persuasive elements of political speech can help social scientists measure political impact \citep{altikriti2016persuasive,hashim2015speech}. We benchmark persuasion strategy and discourse acts classification in \S\ref{subsub:utterance_persuasion} and \S\ref{subsub:discourse_acts}.
\subsection{Psychology} 
\label{subsec:psychology}
As the science of mind and behavior, psychology intersects all other adjacent social sciences in this section. For example, an individual's personality, or their stable patterns of thought and behavior across time, will correlate with their political leaning \citep{gerber2010personality}, social status \citep{anderson2001attains}, and linguistic expression \citep{pennebaker1999linguistic}. The most influential personality modeling benchmark, MyPersonality \citep{kosinski2013private}, is no longer available, but in this work, we evaluate on a representative set of psychological factors down-stream of personality. For example, differences in personality and cognitive processing can impact what people find funny \citep{martin2018psychology} or persuasive \citep{hirsh2012personalized}. These psychological factors then exert influence over a range of social interactions. Humor and politeness \citep{brown1987politeness} are correlated with subjective impressions of psychological distance between speakers \citep{trope2010construal}, while persuasive techniques bind agents in social commitments, with applications in the science of management and organizations. We evaluate on humor, persuasion, and politeness classification in \S\ref{subsub:humor}, \S\ref{subsub:utterance_persuasion}, and \S\ref{subsub:politeness} respectively. We also consider \llm{s} as tools for counseling, mental health and positive psychology in text-based interactions. Specifically, we evaluate on \textit{empathy detection} in online mental health platforms \citep{sharma2020empathy} in \S\ref{subsub:empathy}, \textit{emotional aspect-based summarization} in \S\ref{subsub:covid_et_summarization} ,and a \textit{positive reframing} style-transfer task \citep{ziems2022inducing} based on cognitive behavioral therapy in \S\ref{subsub:positive_reframing}.
\subsection{Sociology} 
\label{subsec:sociology}
Sociologists want to understand the structure of society and how people live collectively in social groups \citep{wardhaugh2021introduction,keuschnigg2018analytical}. By tracing the diffusion and recombination of linguistic, political, and psychological content between actors in a community across time, sociologists can begin to understand social processes at both the micro and macro scale. At the micro scale, there is the computational sociology of power \citep{danescu2012echoes,bramsen2011extracting,prabhakaran2014gender,prabhakaran2012predicting} and social roles \citep{welser2011finding,fazeen2011identification,zhang2007expert,yangwenrose2015,maki2017roles}. \llm{s} can assist sociological research by predicting power relations (\S\ref{subsub:power}) and unhealthy conversations (\S\ref{subsub:conversation_toxicity}). At the macro-scale, there are computational analyses of social norms and conventions \citep{centola2018experimental,bicchieri2005grammar}, information diffusion \citep{leskovec2009meme,tan2014effect,vosoughi2018spread,cheng2016cascades}, emotional contagion \citep{bail2016emotional}, collective behaviors \citep{barbera2015critical}, and social movements \citep{nelson2021cycles,nelson2015political}. Again, \llm{s} can detect constructs like emotion (\S\ref{subsub:emotion}) and the speech of hateful social groups (\S\ref{subsub:utterances_hate_speech}). Furthermore, social movements rely on the diffusion of norms and idiomatic slogans, which carry meaning through figurative language that \llm{s} can decode (\S\ref{subsub:figurative_language}).
\section{Representative CSS Task Selection}
\label{sec:tasks}
While not exhaustive, our task selection is designed to provide a representative survey of the CSS needs in \S\ref{sec:core_methods_css}. \revision{This will allow us to answer Research Questions 1-4, which are pertinent to social science researchers. Thus our work is distinct and complementary to BIG-Bench \citep{srivastava2023beyond} and other efforts to benchmark the logical, physical, and social reasoning capabilities of \llm{s}. Our attention is more carefully focused on the affordances of \llm{s} for social science.}\footnote{To elaborate, BIG-Bench, among its 200 tasks, has some overlap in \textit{figurative language, humor, emotion, empathy}, and \textit{toxicity detection}, but it does not cover \textit{dialect, discourse relations, character tropes, event detection, ideology, misinformation, persuasion, politeness, power relations, semantic change}, or \textit{stance}. BIG-Bench covers few document-level analyses and no conversation-level analysis. We are the first to run extensive experiments to understand patterns of \llm{} performance on tasks critical to social scientists.}\revision{Our tasks are field-specific and come with the field-specific challenges of expert taxonomies, large label spaces, temporal grounding, and domain-specific parsing schemes (see \S\ref{subsec:css_challenges}).}

To help answer RQ3 and 4, we organize this section according to our division of tasks into functional categories based on the unit of text analysis: 10 utterance-level classification tasks (\S\ref{subsec:utterances}), 6 conversation-level tasks (\S\ref{subsec:conversations}), and 4 document-level tasks for the analysis of media (\S\ref{subsec:document}). In addition to these \numClfTasks{} classification tasks, we evaluate \numGenTasks{} generation tasks in \S\ref{subsec:generation} for explaining social science constructs and applying psychological theories to restructure text.
\subsection{Utterance-Level Classification}
\label{subsec:utterances}
An utterance is a unit of communication produced by a single speaker to convey a single subject, which may span multiple sentences \citep{bakhtin2010speech}. CSS researchers can use utterance data to study linguistic phenomena like the syntax of dialect, the semantics of figurative language, or the pragmatics of humor. Utterance-level analysis also reflects human states like emotion and communicative intent, or stable traits like stance and ideology \citep{evans2016machine}. We evaluate LLMs on utterance classification tasks for dialect, hate speech, figurative language, emotion, humor, misinformation, ideology, persuasion, semantic change, and stance classification.

\subsubsection{Dialect Features}
\label{subsub:dialect_feature_detection}
Linguistic feature detection is critical to the study of dialects \citep{eisenstein2011discovering} and ideolects \citep{zhu2021idiosyncratic}, with numerous applications in sociolinguistics, education, and the sociology of class and community membership (see \S\ref{subsec:linguistics}). These features can be used to study the sociolinguistics of language change \citep{kulkarni2015statistically,hamilton2016diachronic} or the linguistic biases in educational assessments \citep{craig2002oral} and online moderation \citep{sap2019risk}. The utterance is an appropriate level of analysis here because syntactic and morphological features are all defined on subtrees of the sentence node  \citep{ziems-etal-2023-multi,eisenstein2023md3}.

We evaluate on the Indian English dialect feature detection task of \citet{demszky2020learning} because this is one of the only available datasets to be hand-labeled by a domain expert. Additionally, Indian English is the most widely-spoken low-resource variety of English, so the domain is representative. The task is to map utterances to a set of 22 grammatical features: i.e., a lack of inversion in $wh$-questions, the omission of copula \textit{be}, or features related to tense and aspect like the \textit{habitual progressive}, found in Indian varieties of English. \revision{For example, the sentence ``\textit{Two years I stayed alone}'' exemplifies \textit{Preposition Omission}.}

\subsubsection{Emotions}
\label{subsub:emotion}
Emotion detection, the cornerstone of affective computing \citep{picard2000affective}, is highly relevant to psychology and political science, among other disciplines, since stable emotional patterns in-part define an individual's personality, and targeted emotions outline the political stances she has. Additional application domains  for the task include emotional contagion \citep{bail2016emotional} and human factors behind economic markets \citep{bollen2011twitter,nguyen2015topic}.

Expert-labeled emotion detection datasets are not common. We evaluate emotion detection with weakly labeled Twitter data from \citet{saravia2018carer}, which uses \citeauthor{plutchik1980general}'s 8 emotional categories: \textit{anger, anticipation, disgust, fear, joy, sadness,
surprise,} and \textit{trust}. \revision{For example, the following sentence would express \textit{fear}:}
\begin{quote}
    \revision{I started the steroids on Saturday and I had some really bad side effects, like my eyes started feeling weird.}
\end{quote}
Plutchik's model is one of the three most recognized discrete emotion models, and it is also used in our later Emotion Summarization Task (\S\ref{subsub:covid_et_summarization}).

\subsubsection{Figurative Language}
\label{subsub:figurative_language}
Figurative expressions are where the speaker meaning differs from the utterance's literal meaning. Recognizing figurative language is a first step in understanding literary content \citep{jacobs2018makes} and political texts \citep{huguet-cabot-etal-2020-pragmatics}, detecting hate speech \citep{lemmens-etal-2021-improving} and identifying mental health self-disclosure \citep{iyer2019figurative}.

We use the FLUTE \citep{chakrabarty2022flute} benchmark because it is, at this time, the most comprehensive, with examples from many prior datasets \citep{chakrabarty2021figurative, srivastava2023beyond,stowe-etal-2022-impli}. FLUTE contains 9k premise sentences, each paired to a hypothesis with figurative language: 
\begin{quote}
    \revision{premise: I said, work independently and come up with some plans. 
\newline\newline
hypothesis: I said, put your heads together and come up with some plans.}
\end{quote}

The classification task is to recognize whether the figurative sentence contains (1) \textit{sarcasm} \citep{joshi2017automatic}, (2) \textit{simile} \citep{niculae-danescu-niculescu-mizil-2014-brighter}, (3) \textit{metaphor} \citep{gao2018neural}, or (4) an \textit{idiom} \citep{jochim-etal-2018-slide}. \revision{In the above example, the hypothesis contains the idiom ``put your heads together.''}

\subsubsection{Hate Speech}
\label{subsub:utterances_hate_speech}
Hate speech is language that disparages a person or group on the basis of protected characteristics like race. Beyond the societal importance of detecting and mitigating hate speech, this is a category of language that is salient to many social scientists. By not only detecting, but also systematically understanding hate speech, political scientists can track the rise of hateful ideologies, and sociologists can understand how these hateful ideas diffuse through a network and influence social movements. 

Thus we evaluate on the more nuanced task of fine-grained hate speech taxonomy classification from Latent Hatred \citep{elsherief2021latent}. This task requires models to infer a subtle social taxonomy from the coded or indirect speech of U.S. hate groups. Utterances should be classified with one of six domain-specific categories: \textit{incitement to violence, inferiority language, irony, stereotypes and misinformation, threatening and intimidation language}, and \textit{white grievance}. \revision{For example, the following sentence contains \textit{white grievance}: ``jewish harvard professor noel ignatiev wants to abolish the white race.''}

\subsubsection{Humor}
\label{subsub:humor}
Humor is a rhetorical \citep{markiewicz1974effects} and literary device \citep{kuipers2009humor} that modulates social distance and trust \citep{sherman1988humor,graham1995involvement,kim2016supervisor}. However, different audiences may perceive the same joke differently. In the study of sociocultural variation, communication, and bonding, humor detection will be of prime interest to sociologists and social psychologists, as well as to literary theorists and historians. Computational social scientists have effectively detected punchlines \citep{mihalcea-strapparava-2005-making, ofer-shahaf-2022-cards} and predicted audience laughter \citep{chen-soo-2018-humor}, demonstrating the computational tractability of the domain.

Our evaluation uses a popular dataset from \citet{weller2019humor} to focus on binary humor detection across a wide range of joke sources, from Reddit's \texttt{r/Jokes}, a \textit{Pun of The Day} website, and a set of short jokes from Kaggle, summing to $\sim$16K jokes. 

\subsubsection{Ideology}
\label{subsub:utterance_ideology}
A speaker's subtle decisions in word choice and diction can betray their beliefs and the political environment to which they belong \citep{jelveh2014detecting}. While political scientists care most about identifying the underlying ideologies and partisan organizations behind these actors (\S\ref{subsec:polisci}), sociolinguists can study the correlation between language and social factors.

We evaluate ideology detection on the Ideological Books Corpus \citep{gross2013testing} from \citet{iyyer2014political}, which contains 2,025 liberal sentences, 1,701 conservative sentences, and 600 neutral sentences. The corpus was designed to disentangle a speaker's overall partisanship from the particular ideological beliefs that are reflected in an individual utterance. Thus labels reflect \textit{perceived} ideology according to annotators and not the speaker's ground truth partisan affiliation. \revision{For example, one sentence associated with a strongly \textit{conservative} ideology is: ``\textit{the feminist movement, with its mockery of marriage and demands for absolute sexual freedom... was a frontal assault on the meritocracy and the traditional family.}''}

\subsubsection{Misinformation}
\label{subsub:misinfo_classification}
Misinformation is both a political and social concern as it can jeopardize democratic elections, public health, and economic markets. The effort to combat misinformation is multi-disciplinary \citep{lazer2018science}, and it depends on reliable misinformation detection tools.

We evaluate on the Misinfo Reaction Frames corpus \citep{gabriel2022misinfo}, a dataset of 25k news headlines with fact checked labels for the accuracy of the related news articles about COVID-19, climate change, or cancer. Models perform binary misinformation classification on news article headlines alone, which the authors found was a tractable task for fine-tuned models. \revision{For example, an article with the headline ``\textit{White House Ousts Top Climate Change Official}'' is marked as likely to contain misinformation.}

\subsubsection{Persuasion}
\label{subsub:utterance_persuasion}
Persuasion is the art of changing or reinforcing the beliefs of others. Understanding persuasive strategies is central to behavioral economics and the psychology of advertising and propaganda \citep{martino2020survey}. Utterances are a natural unit for the analysis of individual persuasive strategies, which may be combined in dialogue for an overall persuasive effect (c.f. \S\ref{subsub:conversation_persuasion}).

While multi-modal persuasion detection tasks exist, we focus on the popular text-based persuasion dataset, Random Acts of Pizza \citep[RAOP;][]{althoff2014ask}, where Reddit users attempt to convince community members to give them free food. This dataset was labeled by \citet{yang2019let} with a fine-grained persuasive strategy taxonomy based on \citet{cialdini2003influence} that includes \textit{Evidence, Impact, Politeness, Reciprocity, Scarcity}, and \textit{Emotion}. 
The task objective is to classify utterance-level RAOP requests according to this 6-class taxonomy. \revision{An example of an \textit{Evidence} sentence is ``\textit{There is a Pizza Hut and a Dominos near me}," since it provides concrete facts relevant to the request. An example of Scarcity is ``\textit{I haven't had a meal in two days}.''}

\subsubsection{Stance}
\label{subsub:utterance_stance}
Although stance detection can be formalized in different ways, the most common task design is for models to determine whether a text's author is in favor of a target view, against the target, or neither. With this formulation, sociologists can understand consensus and disagreement in social groups, psychologists can measure interpersonal attachments, network scientists can build signed social graphs, political scientists can track the views of a voter base or the policies of candidates, historians can plot shifting opinions, and digital humanities researchers can quickly summarize narratives via character intentions and goals.

We evaluate stance detection on the earliest and most established SemEval-2016 Stance Dataset \citep{mohammad2016semeval}, which contains 1,250 tweets and their associated stance towards six topics: \textit{atheism, climate change, the feminist movement, Hillary Clinton, Donald Trump} and the \textit{legalization of abortion.} Stance is given as \textit{favor}, \textit{against}, or \textit{none}. \revision{For zero-shot experiments, we use the test set where the target is Donald Trump for all evaluations. An example tweet \textit{against} Donald Trump is as follows:}
\begin{quote}
    \revision{@realDonaldTrump needs to learn when to stop talking. You are making it worse Donald... so much worse.}
\end{quote}

\subsubsection{Semantic Change}
\label{subsub:semantic_change}
In addition to its more stable features, researchers can plot the change of language over time for a fixed community. Semantic change detection can serve as a proxy measure for the spread and change of culture \citep{kirby2007innateness}, both on the internet~\citep{eisenstein2012mapping, eisenstein2014diffusion} and in historical archives~\citep{mihalcea-nastase-2012-word, kim-etal-2014-temporal, kulkarni2015statistically, rudolph2018dynamic}\footnote{Additional works in this area can be found under the \href{https://www.aclweb.org/portal/content/3rd-workshop-computational-approaches-historical-language-change}{Workshop on Computational Approaches to Historical Language Change}}.

We evaluate \llm{s} as binary word-sense discriminators using the popular Temporal Word-in-Context benchmark~\citep[TempoWiC;][]{wic}. TempoWiC measures the core capability of drawing discrete boundaries between word-level semantics. Given two sentences with the same lexeme, the task is binary classification with positive indicating both sentences use the same sense of the word and negative indicating different senses of the word. \revision{For example, consider the different senses of the word `impostor' in the following texts.}
\begin{quote}
    \revision{text1: Having a rough start to my doctorate program in both the student and teacher roles and feel down and ashamed. I spoke to faculty and know how to move forward, but while they believe in me I find it hard to believe in myself. How do you fight \textbf{impostor} syndrome @AcademicChatter\newline\newline
text2: laughed so hard running from \textbf{impostor} friend around the lab table that I gave myself an headache lmao what a good day}
\end{quote}
A perfect classifier for this task can be used to cluster all usage of a surface-form into sense groups using pairwise comparison.

\subsection{Conversation-Level Classification}
\label{subsec:conversations}
Conversations are multi-party exchanges of utterances. They are critical units for analysis in the social sciences \citep{hutchby2008conversation,silverman1998harvey,sacks1992lectures}, since they richly reflect social \textit{relationships} \citep{evans2016machine} --- a key factor that was missing in utterance-level analysis. Sociological frameworks like ethnomethodology \citep{garfinkel2016studies} focus particularly on conversations. The tasks in this section are drawn largely from the ConvoKit toolkit of \citet{chang2020convokit}.

\subsubsection{Discourse Acts}
\label{subsub:discourse_acts} 
Discourse acts are the building blocks of conversations and are thus relevant to conversation analysis in sociology, genre analysis in literature, pragmatics, and ethnographic studies of speech communities (see \citeauthor{paltridge2000making} for example). Some popular discourse act taxonomies like DAMSL \citep{stolcke2000dialogue} and DiAML \citep{bunt2010towards} can have as many as 40 categories, tailored to spoken communication. We use \citet{zhang2017characterizing}'s simpler and more focused 9-class taxonomy since it was designed to cover \textit{online} text conversations---the focus of CSS research. The taxonomy includes \textit{questions, answers, elaborations, announcements, appreciation, agreements, disagreements, negative reactions}, and \textit{humor}.

We evaluate on the Coarse Discourse Sequence Corpus \citep{zhang2017characterizing}. The model input is a comment from a Reddit thread, along with the utterance to which the comment is responding. For example, 
\begin{quote}
    \revision{[userABC]: So i went thinking to myself this fine day \"hey lets check out Levetihan\" and then i found out that this DLC does not appear in my Origin store...
    \newline \newline
    [userXYZ]: Did you go into ME3 game and access \"downloadable content\"?}
\end{quote}
The expected output is the category from the above 9-class taxonomy which best describes the comment's speech act: \textit{Question} in the above example.. Since \textit{Announcements} and \textit{Negative reactions} have fewer than 10 examples total in the dataset, they are omitted from our evaluation along with the catch-all \textit{Other} category.

\subsubsection{Empathy}
\label{subsub:empathy} Since the early days of internet access, users have looked to internet communities for support~\citep{preece1998empathic}. Thus web communities can provide CSS researchers with empathetic communication data in naturalistic settings~\citep{pfeil2007patterns, sharma2020empathy}. By better understanding community-specific affordances \citep{zhou-jurgens-2020-condolence} and the most common triggers for empathetic responses~\citep{buechel-etal-2018-modeling, omitaomu2022empathic}, CSS can reciprocally inform the design of empathetic communities \citep{coulton2014designing, taylor2019accountability}, as well as community-specific tools like counseling dialogue systems~\citep{sharma2021towards, ma2020survey}.

Understanding is the first step towards building more effective online mental health resources, and this motivates our evaluation on the TalkLife dataset of~\citet{sharma2020empathy}, a clinically-motivated empathy detection dataset. The paper's EPITOME measures empathy using 
a multi-stage labeling scheme. First, a listener communicates an \textit{Emotional Reaction} to describe how the seeker's disclosure makes the listener feel. Then the listener offers an \textit{Interpretation} of the emotions the seeker is experiencing. Finally, the listener moves into \textit{Exploration}, or the pursuit of further information to better understand the seeker's situation. Clinical psychologists labeled the listener's effectiveness at each stage of a listener's top-level reply. Here we focus on \textit{Exploration}, as prior work has shown open-questions to be especially effective for peer-support~\citep{shah2022modeling}. Given a seeker's post and a top-level listener's reply, we classify whether the listener offered: \textit{Strong Exploration} (specific questions about the seekers situation), \textit{Weak Exploration} (general questions), or \textit{No Exploration.} Consider the example conversation:
\begin{quote}
    \revision{Seeker: I spent today either staring blankly at a computer screen or my phone. Was too hurt to do anything today, really.\newline\newline
Response: I wish I even had the will to play games. For me it's excessive daydreaming.}
\end{quote}
\revision{The above response is an example of \textit{No exploration.}}

\subsubsection{Persuasion}
\label{subsub:conversation_persuasion}
In \S\ref{subsub:conversation_persuasion}, we considered utterance-level analysis of fine-grained persuasive strategies. However, social scientists are also interested in the overall persuasive effect that one speaker has on another through sequences of rhetorical strategies in dialogue \citep{shaikh-etal-2020-examining}. Persuasive outcomes are particularly important for the political science of successful campaigns \citep{murphy2004persuasion} and the sociology of idea propagation and social movements \citep{stewart2012persuasion}.

We evaluate our persuasion prediction task on the Winning Arguments Corpus \citep{Tan:2016:WAI:2872427.2883081}, which contains 3,051 conversations from \texttt{r/ChangeMyView} in which the persuader tries to convince the persuadee to change their mind. Models receive as input the reply thread (starting from a top-level comment) and perform binary prediction on whether the persuadee awarded the persuader a `delta' for a successful argument: \textsl{If you were the original poster, would this reply convince you?} Consider this example of an unsuccessful argument by UserA below:
\begin{quote}
    \revision{[UserA]: ``Right on red'', when it's allowed, is primarily because you're going from the innermost lane TO the intersecting innermost lane... this part of the state is infamous for a**hat drivers blocking people from turning, changing lanes, etc... could you imagine the chaos? :D\newline\newline
    [UserB]: Er, the image I posted, I found on Google images. It may triple the number of lanes to be concerned about, but one of them shouldn't usually be a problem if people stay in their lanes. And the other two, you still have to look in the same direction."}
\end{quote}

\subsubsection{Power and Status}
\label{subsub:power}
Sociologists, political scientists, and online communities researchers are interested in understanding hierarchical organizations, social roles, and power relationships. Power is related to control of the conversation \citep{prabhakaran2014gender} and power dynamics shape both behavior and communication. Specifically, text analysis can uncover power relationships in the degree to which one speaker accommodates to the linguistic style of another \citep{danescu2012echoes}. We anticipate that this task is tractable for \llm{s}.

We evaluate on the Wikipedia Talk Pages dataset from \citet{danescu2012echoes}. Conversations are drawn from the debate forums regarding Wikipedia edit histories, and power is a binary label describing whether or not the Wikipedia editor is an administrator. All models are given an editor's entire comment history from the Talk Pages, and the objective is binary classification. \revision{In the following example, EditorA uses a high degree of politeness and hedging langauge, which indicates that he is not in a position of power:}
\begin{quote}
    \revision{[EditorA]: That's odd.  Somehow, I came across one of that user's edits, though I believe it was on recent changes.  As you  can see, most of the older edits are vandalism, but I guess due to the time that wouldn't warrant much of a block.  I don't know how I happened to come across that since it's so old.\newline\newline
    [EditorA]: That could be the case.  I've seen a few of those tonight.}
\end{quote}

\subsubsection{Toxicity Prediction}
\label{subsub:conversation_toxicity} Toxicity is a major area of social research in online communities, as online disinhibition~\citep{suler2005contemporary} makes antisocial behaviour especially prevalent~\citep{cheng2015antisocial}. Predictive models can be used to understand the early signs of later toxicity~\citep{cheng2017anyone} for downstream causal analysis on the evolution of toxicity~\citep{mathew2020hate} and the effectiveness of intervention methods~\citep{kwak2015exploring}. Even without interpretable features, a predictive system can serve causal methods as a propensity score.

Using the Conversations Gone Awry corpus~\citep{zhang2018conversations}, we investigate whether \llm{s} can predict future toxicity from early cues. As context, the model takes the first two messages in a conversation between Wikipedia users. The model should make a binary prediction whether or not the Wikipedia conversation will contain toxic language at any later stage. For example:
\begin{quote}
\revision{[UserA]: I have removed recent edition  of pappe to the lead though Pappe view might notable currently without attribution and proper context of other views it WP:NPOV  violation.\newline\newline
[UserB]: In fact, Pappe is already mentioned twice in the proper place.}
\end{quote}
\revision{The conversation above contains overt confrontation that will later devolve into toxicity.}
\subsubsection{Politeness}
\label{subsub:politeness} Before overt toxicity is evident in a community, researchers can measure its health and stability according to members' adherence to politeness norms. Polite members can help communities grow and retain other valuable members~\citep{burke2008mind}, while rampant impoliteness in a community can foreshadow impending toxicity~\citep{andersson1999tit}. Text-based politeness measures also reflect other societal factors that we explore in this work, like gender bias~\citep[\S\ref{subsub:utterances_hate_speech}]{herring1994politeness, ortu2016emotional}, power inequality~\citep[\S\ref{subsub:power}]{danescu-niculescu-mizil-etal-2013-computational}, and persuasion~\citep[\S\ref{subsub:utterance_persuasion}]{shaikh-etal-2020-examining}. 

We evaluate on the Stanford Politeness Corpus~\citep{danescu-niculescu-mizil-etal-2013-computational}. The dataset is foundational in the computational study of politeness and its relation to other social dynamics. The corpus contains requests made by one Wikipedia contributor to another. For example,
\begin{quote}
    \revision{I am looking for help improving the dermatology content on Wikipedia. Would you be willing to help, or do you have any friends interested...}
\end{quote}
Each request is classified into one of three categories, \textit{Polite, Neutral}, or \textit{Impolite}, according to Mechanical Turk annotators' interpretation of workplace norms \revision{(the example above is \textit{Polite})}. High zero-shot performance on this task will strongly indicate a model's broader ability to recognize conversational social norms.

\subsection{Document-Level Classification}
\label{subsec:document}
Documents provide a complementary view for social science. Like conversations, documents can encode sequences of ideas or temporal events, as well as interpersonal relationships not present in isolated utterances. Unlike the dyadic communication of a conversation, a document can be analyzed under a unified narrative \citep{piper2021narrative}. Thus for our purposes, a document is a collection of utterances that form a single \textit{narrative}. 
Our document-level classification tasks cluster around \textit{media}, which has been the subject of content analysis in the social sciences since the time of Max Weber in 1910. In this section, we focus on computational tools for content analysis \citep{berelson1952content} to code media documents for their underlying \textit{ideological} content (\S\ref{subsub:media_ideology}), the
\textit{events} they portray (\S\ref{subsub:event_detection}, \S\ref{subsub:event_argument_extraction}), as well as the \textit{agents} involved and the specific \textit{roles} or character tropes they exhibit (\S\ref{subsub:role_tagging}).

\subsubsection{Event Detection}
\label{subsub:event_detection}
Following a massive effort to digitize critical documents, social scientists depend on event extraction to automatically code and organize these documents into smaller and more manageable units for analysis. Events are the ``building blocks'' from which historians construct theories about the past \citep{sprugnoli2019novel}; they are the backbone of narrative structure \citep{chambers-jurafsky-2008-unsupervised}. Event detection is the first step in the event extraction pipeline.
Hippocorpus \citep{sap2020recollection} is a resource of 6,854 stories that were collected from crowdworkers and tagged for sentence-level events \citep{sap2022quantifying} . Events can be further classified into minor or major events, as well as expected or unexpected. We evaluate on the simplest task: binary event classification at the sentence level. For example:
\begin{quote}
    \revision{A: Four months ago, I had a big family reunion.\newline
B: We haven't had one in over 20 years.\newline
C: This was a very exciting event.\newline
D: I saw my Grandma who said I liked great as ever.}
\end{quote}
The above lines A and D denote new events. 
\subsubsection{Event Argument Extraction}
\label{subsub:event_argument_extraction}
Where event detection was concerned with identifying event triggers, event argument extraction is the task of filling out an event template according to a predefined ontology, identifying all related concepts like participants in the event, and parsing their roles. Historians, political scientists, and sociologists can use such tools to extract arguments from sociopolitical events in the news and historical text documents, and to understand social movements \citep{hurriyetouglu2021challenges}. Economists can use event argument extraction to measure socioeconomic indicators like the unemployment rate, market volatility, and economic policy uncertainty \citep{min-zhao-2019-measure}. Event argument extraction is also a key feature of narrative analysis \citep{sims2019literary}, as well as in the wider domains of legal studies \citep{shen2020hierarchical}, public health \citep{jenhani2016hybrid}, and policy.

WikiEvents \citep{li2021document} is a document-level event extraction benchmark for news articles that were linked from English Wikipedia articles. WikiEvents uses DARPA's KAIROS ontology with 67 event types in a three-level hierarchy. For example, the \texttt{Movement.Transportation} event has the agentless \texttt{Motion} subcategory and an agentive \texttt{Bringing} subcategory. Both include a \texttt{Passenger}, \texttt{Vehicle}, \texttt{Origin}, and \texttt{Destination} argument, but only the agentive \texttt{Bringing} has a \texttt{Transporter} agent. KAIROS's event argument ontology is richer and more versatile than the commonly used ACE ontology, which only has 33 types of events. \revision{An example of this task is to take the following document}
\begin{quote}
    \revision{The Taliban <tgr>killed <tgr>more than 100 members of the Afghan security forces inside a military compound in central Maidan Wardak province on Monda...}
\end{quote}
\revision{and produce the following structured output}
\begin{lstlisting}
    {'Victim': 'members', 'Place': 'undefined', 'Killer': 'The Taliban', 'MedicalIssue': 'undefined'}
\end{lstlisting}

\subsubsection{Ideology}
\label{subsub:media_ideology} CSS is extremely useful for understanding and quantifying real and perceived political differences. For a variety of specific phenomena~\citep{amber2013identifying, baly-etal-2018-predicting, roy2020weakly, luo-etal-2020-detecting, ziems2021protect}, this takes the form of gathering articles from across the political spectrum, processing each one further for a phenomenon of interest, and evaluating the relative differences for the articles from different ideological groups. The first step in such studies is to separate articles according to the overarching political ideology they represent.

We evaluate ideology detection on the Article Bias Corpus from \citet{baly2020we}, which collects a set of articles from media sources covering the United States of America and labels them according to Left, Right, and Centrist political bias. Unlike the task of utterance-level ideology prediction (\S\ref{subsub:utterance_ideology}), this task provides an entire news article as context. This tests the ability of the model to understand the relationship that a \textit{sequence} of stances taken across an entire article might have with political leaning. \revision{For example, one Left-leaning article in this dataset contains the strongly-indicative phrase: ``\textit{it was hard not to think about the insularity and cossetting the super-wealthy enjoy,}'' and then goes on to talk at length about former LA Clippers owner Donald Sterling. In other articles, political views are more diffuse and less starkly concentrated into particular phrases or slogans. Still, each article must be classified into exactly one of the three ideological categories above.}

\subsubsection{Roles and Tropes}
\label{subsub:role_tagging}
Social roles are defined by expectations for behavior, based on social interaction patterns \citep{yang2019seekers}. Similarly, personas are simplified models of personality \citep{grudin2006personas}, like a trope that a character identifies within a movie. The ability to infer social roles and personas from text has immediate applications in the psychology of personality, the sociology of group dynamics, and the study of agents in literature and film. These insights can help us understand stereotypical biases and representational harms in media \citep{blodgett-etal-2020-language}.
Downstream applications also include narrative psychology \citep{murray2015narrative}, economics, political polarization, and mental health \citep{piper2021narrative}.

Others have considered character role labeling for narratives \citep{jahan2021inducing} and news media \citep{gomez2018hero}. We evaluate this task with the CMU Movie Corpus dataset from \citet{bamman2013learning} as it was extended and modified by \citet{chu2018learning} to include character trope labels and IMBD character quotes. The \textit{character trope classification} task involves identifying from a character's quotes alone which of 72 movie tropes that characters identity best fits; e.g., the \textit{coward} or the \textit{casanova}. The following example quotations are from an \textit{absent-minded professor}:
\begin{quote}
    \revision{Now, THIS makes any fabric instantly impervious. Dirt proof, stain proof... Ouch! And bullet proof! It's still not perfected yet! It's hell on the dry-cleaning bill.\newline\newline
    This baby is the ultimate corrosive. I call it - DON'T TOUCH IT! - I call it ``hydrochloricdioxynucleocarbonium''. Well, the name needs work. But it'll eat through a Buick! OR -}
\end{quote}

\subsection{Generation Tasks}
\label{subsec:generation}
Regarding RQ4 \textbf{Functionality}, we want to understand whether \llm{s} are best suited to classify taxonomic social science constructs from text, or whether these models are equally if not better suited for generative explanations and summaries. This section describes our natural language generation tasks, where \llm{s} might be used to summarize relevant aspects (\S\ref{subsub:covid_et_summarization}), elucidate the hidden social meaning behind a text (\S\ref{subsub:figurative_langauge_explanation} --- \S\ref{subsub:sbic_explanation}) or implement social theory by stylistically restructuring an utterance (\S\ref{subsub:positive_reframing}).

\subsubsection{Emotion-Specific Summarization}
\label{subsub:covid_et_summarization}
\revision{Prior work has already demonstrated \llm{s'} skill at generic summarization tasks \citep{goyal2022news,qin2023chatgpt}. Here, we consider a more domain-specific task, \textit{aspect-based summarization}. The key idea of aspect-based summarization is that different elements of a document will be relevant to different users. This is especially true for social scientists and other domain specialists. For example, scientists who study population-level emotional responses to crisis events will need to know not only which emotions are represented in text (i.e., emotion detection; \S\ref{subsub:emotion}), but also, in brief, what specific experiences triggered such emotions. The scientist may not have highly focused queries like in QA tasks, but this use still demands summaries about broad subtopics or themes  \citep{ahuja2022aspectnews}.}

\revision{We use the \textsc{CovidET} dataset of \citet{zhan2022you}, which contains 1,883 Reddit posts from the COVID-19 pandemic. Given a post and one of \citeauthor{plutchik1980general}'s target emotions, the task is to summarize from the post what triggered the author to feel the target emotion.}

\subsubsection{Figurative Language Explanation}
\label{subsub:figurative_langauge_explanation}
Our interests in figurative language are covered in \S\ref{subsub:figurative_language}, where we introduce the FLUTE dataset.
FLUTE contains 9k (literal, figurative) sentence pairs with either entailed or contradictory meanings. The goal of the explanation task is to generate a sentence to explain the entailment or contradiction. For example, the figurative sentence ``she absorbed the knowledge'' entails the literal sentence ``she mentally assimilated the knowledge'' under the following explanation: ``to absorb something is to take it in and make it part of yourself.''

\subsubsection{Implied Misinformation Explanation}
\label{subsub:misinfo_explanation}
Both scientific understanding and real-world intervention strategies depend on more than black-box classification. This motivates our implied statement generation task, \revision{which is specified in the Misinfo Reaction Frames corpus \citep{gabriel2022misinfo} as covered in \S\ref{subsub:misinfo_classification}}. Models take the headline of a news article and generate the underlying meaning of the headline in plain English. This is called the \textit{writer's intent.} Consider, for example, the misleading headline, ``\textit{Wearing a face mask to slow the spread of COVID-19 could cause Legionnaires’ disease.}'' Here, the annotator wrote that the writer's intent was to say ``\textit{wearing masks is dangerous; people shouldn't wear masks.}''

\subsubsection{Social Bias Inference}
\label{subsub:sbic_explanation}
While hate speech detection focuses on the overall harmfulness of an utterance, specific types of hate speech are targeted towards a demographic subgroup. To this end, the Social Bias Inference Corpus~\citep[SBIC;][]{sap2020social} consists of 34K inferences, where hate speech is annotated with free-text explanations. Importantly, explanations highlight \textit{why} a specific subgroup is targeted. For example, the sentence \textit{``We shouldn’t lower our standards just to hire more women.''} implies that \textit{``women are less qualified.''} To model these explanations, \citet{sap2020social} treat the task as a standard conditional generation problem. We mirror this setup to evaluate LLMs.

\subsubsection{Positive Reframing}
\label{subsub:positive_reframing}
NLP can help scale mental health and psychological counseling services by training volunteer listeners and teaching individuals the techniques of cognitive behavioral therapy (CBT; \citeauthor{rothbaum2000cognitive}, \citeyear{rothbaum2000cognitive}), which is used to address mental filters and biases that perpetuate anxiety and depression. Positive reframing is a sequence-to-sequence task which translates a distorted negative utterance into a complementary positive viewpoint using CBT strategies without contradicting the original speaker meaning. Take the example source sentence:
\begin{quote}
    \revision{Always stressing and thinking about loads of things at once need I take it one at a time overload stressed need to rant.}
\end{quote}
Using the \textit{growth} and \textit{neutralizing} strategies, the author can reframe this thought more positively as follows:
\begin{quote}
    \revision{Loads of things on my mind, I need to make a list, prioritise and work through it all calmly and I will feel much better.}
\end{quote}

\section{Evaluation Methods}

\subsection{Model Selection and Baselines}
\label{subsec:model_selection}
Our goal is to evaluate \llm{s} in \textbf{zero-shot settings through prompt engineering} (\S\ref{subsec:prompt_engineering}) and to identify suitable model architectures, sizes, and pre-training/fine-tuning paradigms for CSS research (RQ 1,2). We choose \textbf{FLAN-T5} \citep{chung2022scaling} as an open-source model with strong zero-shot and few-shot performance. Although it follows a standard T5 encoder-decoder architecture, FLAN's zero-shot performance is due to its instruction fine-tuning over a diverse mixture of sequence to sequence tasks. The added benefit is that FLAN-T5 checkpoints exist at six different sizes ranging from small (80M parameters) to XXL (11B) and UL2 (20B), allowing us to investigate scaling laws. Next, we consider OpenAI's \textbf{GPT-3} \citep{brown2020language,zong2022survey} including \texttt{text-001}, \texttt{text-002} learning with instructions and \texttt{text-003}, which is further learned from human preferences (RLHF) \citep{christiano2023deep} series, and \texttt{gpt-3.5-turbo} \citep{qin2023chatgpt,gilardi2023chatgpt}, which is the conversation-based \llm{} trained through RLHF \citep{christiano2023deep}. \revision{Finally we include \textbf{GPT-4} \citep{gpt4}, which is a multimodal model that, at 1.7 trillion parameters, scales up the GPT-3 architecture by 1000$\times$.}

Traditional supervised fine-tuned models can serve as \textbf{baselines} for each task. These baselines are intended to provide a comparison point for the utility of \llm{s} for CSS, rather than providing a fair methodological comparison between approaches. For classification tasks, we use RoBERTa-large \citep{liu2019roberta} as the backbone model and tune hyperparameters based on F1 score on the validation set. For generation tasks, we use T5-base \citep{raffel2020exploring} as the backbone model and tune hyperparameters based on average BLEU score on the validation set. We use a grid search to find the most suitable hyperparameters including learning rate \texttt{\{5e-6, 1e-5, 2e-5, 5e-5\}}, batch size \texttt{\{4, 8, 16, 32\}} and the number of epochs \texttt{\{1, 2, 3, 4\}}.  Other hyperparameters are set to the defaults defined by the HuggingFace Trainer. We average results across three different random seeds to reduce variance. These baselines will prove competitive in Table~\ref{tab:clf_results}, matching or exceeding the best reported performances from the original publications in \textit{Event Surprisal, Event Argument Extraction}, and the classification of \textit{Emotions, Empathy, Figurative Language, Implicit Hate, Persuasion, Persuasion Strategies, Political Ideology}, and \textit{Semantic Change}. Still, it is important to note that greater performances might be achievable by fine-tuning alternative architectures.

\begin{table*}[t]
    \centering
    \resizebox{\textwidth}{!}{%
    \begin{tabular}{p{0.56\textwidth}|c|p{0.38\textwidth}}
    \toprule
Effective Prompt Guideline & Reference & Guideline Example \\ 
\midrule 
When the answer is categorical, enumerate options as alphabetical \textbf{\textcolor{themered}{multiple-choice}} so that the output is simply the highest-probability token (`A', `B').  &  \citet{hendrycksmeasuring}& \multirow{6}{0.4\textwidth}{\hspace{-2pt}\{\$CONTEXT\}\newline\newline Which of the following describes the above news headline?~\Return\newline
\textbf{\textcolor{themered}{A:}} Misinformation~\Return\newline
\textbf{\textcolor{themered}{B:}} Trustworthy~\Return\newline\{\$CONSTRAINT\}} \\
\cmidrule{1-2}
\textbf{Each option should be separated by a new line} (\Return) to resemble the natural format of online multiple choice questions. More natural prompts will elicit more regular behavior. & \href{https://irmckenzie.co.uk/round1\#:~:text=model\%20should\%20answer.-,Using\%20newlines,-We\%20saw\%20many}{Inverse Scaling Prize} &  \\
\midrule
To promote instruction-following, \textbf{\textcolor{themered}{give instructions \textit{after} the context}} is provided; then \textbf{\textcolor{themepurple}{explicitly state any constraints}}. Recent and repeated text has a greater effect on \llm{} generations due to common attention patterns. 
 & \citet{child2019generating} &  \multirow{7}{0.4\textwidth}{\hspace{-2pt}\{\$CONTEXT\}\newline\textbf{\textcolor{themered}{{\{\$QUESTION\}}}}\newline\newline \textbf{\textcolor{themepurple}{Constraint:}} {Even if you are uncertain}, you \textbf{\textcolor{black}{must pick either ``True'' or ``False''}} without using any other words.}\\
\cmidrule{1-2}
 \textbf{\textcolor{black}{Clarify the expected output}} in the case of uncertainty. Uncertain models may use default phrases like ``\textit{I don't know},'' and clarifying constraints force the model to answer.
  & No Existing Reference &  \\
 \midrule
 When the answer should contain multiple pieces of information, \texttt{\textbf{\textcolor{themered}{request responses in JSON format}}}. This leverages \llm{'s} familiarity with code to provide an output structure that is more easily parsed. & \href{https://github.com/srush/minichain\#typed-prompts}{MiniChain Library} & \multirow{4}{0.4\textwidth}{\hspace{-2pt}\{\$CONTEXT\}\newline\{\$QUESTION\}\newline\newline
\textbf{\textcolor{themered}{\texttt{JSON Output:}}}}\\

    \end{tabular}
    }
    \caption{\textbf{\llm{} Prompting Guidelines} to generate consistent, machine-readable outputs for CSS tasks. These techniques can help solve overgeneralization problems on a constrained codebook, and they can force models to answer questions with inherent uncertainty or offensive language.}
    \label{tab:prompting_guidelines}
\end{table*}

\subsection{Prompt Engineering}
\label{subsec:prompt_engineering}
One strength of current \llm{s} is their ability to be "\emph{programmed}" through natural language instructions~\citep{brown2020language}. This capability has been further improved by training models to explicitly follow these instructions \citep{sanh2021multitask, wang2022super, chung2022scaling, ouyang2022training}. CSS tools can then be developed directly by subject-matter experts using natural language instructions rather than explicit programming language interpretations. In order to evaluate \llm{s}, each task requires a prompt designed to elicit the desired behavior from the model.

\revision{The author who is most familiar with the task writes an initial prompt for it based on the task description and the \textit{design guidelines} below. Then we generate four semantically equivalent perturbations of that prompt using \texttt{gpt-3.5-turbo} as a zero-shot paraphrase model. All results are averaged across \textbf{prompt perturbations} to remove instruction-based variance \citep{perez2021true,zhao2021calibrate}.}

In order to receive consistent, reproducible results we utilize a temperature of zero for all \llm{s}. For models which provide probabilities directly, we constrain decoding to the valid output classes.\footnote{Probability outputs for \href{https://huggingface.co/docs/transformers/main_classes/output\#transformers.modeling_outputs.Seq2SeqLMOutput.logits}{HuggingFace} and \href{https://platform.openai.com/docs/api-reference/completions/create\#completions/create-logprobs}{GPT-3}} For other models, such as \texttt{gpt-3.5-turbo}, we use logit bias to encourage valid outputs during decoding.\footnote{Logit Bias reference for \href{https://platform.openai.com/docs/api-reference/chat/create\#chat/create-logit_bias}{\texttt{gpt-3.5-turbo}}} All other generation parameters are left at the default settings for each model.

\paragraph{CSS Prompt Design Guidelines}
CSS tasks often require models to make inferences about subtext and offensive language. Additionally, CSS codebooks often project complex phenomena into a reduced set of labels.  This raises challenges for the use of \llm{s} which have been refined for general use. When initially exploring \llm{} behavior, we found that models would hedge in the case of uncertainty, refuse to engage with offensive language, and attempt to generalize beyond provided labels. While desirable in a general context, these behaviors make it difficult to use \llm{s} inside a CSS pipeline.

Therefore, we built a set of \revision{guidelines} in Table~\ref{tab:prompting_guidelines}, drawn from both the literature and our own experience with non-CSS tasks as NLP researchers. We explicitly share these guidelines to help CSS practitioners control \llm{s} for their purposes. \revision{There is no claim that the resulting prompts are optimally-engineered for each task; they instead provide reasonable approximations to the kinds of prompts a non-AI expert could design after considering established guidelines. By averaging our results over five prompt pertubations, we reduce the variance in this approximation of standard CSS tool-use.}

\begin{table}[]
    \centering
    \begin{tabular}{l|rr}
        \toprule
        Dataset & Size & Classes \\
        \midrule
        Generation Tasks & 500 & - \\ \midrule
        \multicolumn{3}{c}{\textbf{Utterance Level}} \\[3pt]
        \midrule 
        Dialect & 266 & 23\\
        Persuasion & 399 & 7\\
        Impl. Hate & 498 & 6\\
        Emotion & 498 & 6\\
        Figurative & 500 & 4\\
        Ideology & 498 & 3\\
        Stance & 435 & 3\\
        Humor & 500 & 2\\
        Misinfo & 500 & 2\\
        Semantic Chng & 344 & 2 \\[2pt]
    \end{tabular}
    \quad
    \begin{tabular}{l|rr}
        \toprule
        Dataset & Size & Classes \\
        \midrule
        \multicolumn{3}{c}{\textbf{Conversation Level}} \\
        \midrule 
        Discourse & 497 & 7\\
        Politeness & 498 & 3\\
        Empathy & 498 & 3\\
        Toxicity  & 500 & 2\\
        Power & 500 & 2\\
        Persuasion & 434 & 2\\
        \midrule
        \multicolumn{3}{c}{\textbf{Document Level}} \\
        \midrule 
        Event Arg. & 283 & -- \\
        Evt. Surprisal & 240 & --\\
        Tropes & 114 & 114\\
        Ideology & 498 & 3\\
    \end{tabular}
    \caption{Dataset size and classes count across all selected CSS benchmarks. Datasets are sorted by class count for each task category.}
    \label{tab:dataset_stats}
\end{table}

\subsection{Test Set Construction}
\label{subsec:test_set}

For each task, we evaluate a class-stratified sample of at most 500 instances from the dataset's designated test set. If the designation is missing, we take the class-stratified sample from the entire dataset. Our sampled test sizes and class counts are in Table~\ref{tab:dataset_stats}. All datasets, prompts, and model outputs are released for future comparison and analysis.\footnote{Data Directory of our \href{https://github.com/SALT-NLP/LLMs_for_CSS/tree/main/css_data}{Github Project}}

\subsection{Evaluation Metrics}
\paragraph{Automatic Evaluation}
\revision{Each dataset has a different structure with a different number of labels (see Table~\ref{tab:dataset_stats}), so the use of accuracy is not the most informative metric. Instead, we use compute the macro F1 score for all classification and structured parsing tasks and average over the 5 prompt perturbations.} Since we mapped the label space for each task to an alphabetical list of candidate options and set the logit bias to favor these options (\S\ref{subsec:prompt_engineering}), evaluation scripts use straightforward string-matching.

For high-variation domains like our generation tasks, on the other hand, word-overlap-based machine translation metrics like BLEU~\citep{post-2018-call} are expected to have low correlation with human quality judgments \citep{liu-etal-2016-evaluate}. Here, even embedding-similarity metrics like BERTScore~\citep{zhangbertscore} may be insufficient \citep{novikova-etal-2017-need}. Manual inspection revealed high-quality generation outputs, but the BLEURT~\citep{sellam2020bleurt} score reported zero semantic overlap, and variation in BLEU and BERTScores failed to follow any discernible patters with regards to model preference or scaling laws that we observed by manual inspection. For generation tasks, human evaluation is strongly preferable \citep{santhanam2019towards}, and we describe human evaluation in the following paragraphs.

\paragraph{Human Scoring Evaluation}
\label{subsub:human_scoring_method}
\revision{To get a sense of the generation quality for each task, we recruit a domain expert to blindly evaluate 100-400 model outputs. Evaluations are on 1-5 Likert scales for 4 standard metrics from the NLG literature. Following \citet{fabbri2021summeval}, we define these metrics as follows:}
\begin{itemize}
    \item \textbf{Faithfulness:} \textit{The generation is consistent with the source document and with the definition of the task.}
    \item \textbf{Coherence:} \textit{The generation is well-structured and well-organized. The generation is not just a heap of unrelated information, but forms a coherent body of information about a topic.} \citep{dang2005overview}
    \item \textbf{Relevance:} \textit{The generation includes only important information from the source document; no redundancies or excess information.}
    \item \textbf{Fluency:} \textit{The generation has no formatting problems, capitalization errors or obviously ungrammatical sentences (e.g., fragments, missing components) that make the text difficult to read.}
\end{itemize} 

\revision{All experts are recruited and paid through the Upwork platform. Annotator backgrounds and expertise are summarized in the bottom right pane of Table~\ref{tab:generation_finegrained_results}. For \textsc{CovidET} summarization, we recruit a former CDC health communication specialist with a B.S. in Public Health and an M.S. in Health Education. For \textit{Misinformation}, we enlist a Public Policy Graduate Student with a B.A. in Political Science. For \textit{Figurative Language}, we hire a former writing expert at Grammarly with an M.F.A. For \textit{Social Bias}, we find a Graduate Student with a B.S. in Journalism. And for \textit{Positive Reframing}, we hire a Nurse in Clinical Behavioral Health with a B.A. in Psychology.}

\paragraph{Human Ranking Evaluation}
\label{subsub:human_eval_method}

Instead of scoring or rating target generations on a standard Likert scale, annotators can also rank the model explanations in terms of their \textit{faithfulness} at describing the target construct. The ranking-style evaluation can be less variable than scoring for generation tasks \citep{harzing2009rating,belz2010comparing}. We will use Social Bias Frames as an example to illustrate the general setup for Human Ranking Evaluations. Here, the annotator reviews a \textit{hateful message} and an associated \textit{hate target}. Then they review four \textit{Implied Statements} generated by one of the OpenAI models or pulled from the SBIC's gold human annotations. They are asked to rank these statements from best to worst according to how accurate the \textit{implied statement} is at describing the hidden message from the \textit{hateful message}. In this forced-choice ranking scheme, ties are not allowed, but we use a unanimous vote to determine when a given model outranks human performance. Unanimous vote flattens the variance for explanations of similar quality and reflects only significant differences in quality. 

Pilot annotation proved that crowdworker evaluations can exhibit high variance and instability due to cultural and individual differences, as well as different interpretations of the task. Thus the authors served as blind annotators for this evaluation. Two authors evaluated each task and unanimous voting determined the reported metrics.

\section{Classification Results}
\label{subsec:clf_results}

\begin{table*}[]
    \centering
    \resizebox{\textwidth}{!}{%
    \def\arraystretch{1.15}
    \setlength{\tabcolsep}{4pt}
    \begin{tabular}{l|rrrrrrrrrrrrrrrr}
         \toprule 
          \multirow{3}{*}{\backslashbox[19mm]{\textbf{Data}}{\textbf{Model}}}
          & \multicolumn{2}{c}{Baselines} &  \multicolumn{5}{c}{FLAN-T5} & \multicolumn{1}{c}{FLAN} & \multicolumn{4}{c}{text-001} & \multicolumn{1}{c}{text-002} & \multicolumn{1}{c}{text-003} & \multicolumn{2}{c}{Chat}\\
         \cmidrule(lr){2-3} \cmidrule(lr){4-8} \cmidrule(lr){9-9} \cmidrule(lr){10-13} \cmidrule(lr){14-14} \cmidrule(lr){15-15} \cmidrule(lr){16-17}
         & Rand & Finetune & Small & Base & Large & XL & XXL & UL2 & Ada & Babb. & Curie & Dav. & Davinci & Davinci & GPT3.5 & GPT4  \\
         \midrule
 \multicolumn{16}{c}{\textbf{Utterance Level Tasks}} \\
 \midrule
         Dialect    &       3.3 &     3.0 &                     0.2 &                    4.5 &                    23.4 &                 24.8 &                  30.3 &               \valbest{32.9} &             0.5 &                 0.5 &               1.2 &                 9.1 &                17.1 &                14.7 &       11.7 &     23.2 \\
Emotion    &      16.7 &    71.6 &                    19.8 &                   63.8 &                    69.7 &                 65.7 &                  66.2 &               \valbest{70.8} &             6.4 &                 4.9 &               6.6 &                19.7 &                36.8 &                44.0 &       47.1 &     50.6 \\
Figurative &      25.0 &    99.2 &                    16.6 &                   23.2 &                    18.0 &                 32.2 &                  53.2 &               \valbest{62.3} &            10.0 &                15.2 &              10.0 &                19.4 &                45.6 &                57.8 &       48.6 &     17.5 \\
Humor      &      49.5 &    73.1 &                    51.8 &                   37.1 &                    54.9 &                 56.9 &                  29.9 &               56.8 &            38.7 &                33.3 &              34.7 &                29.2 &                29.7 &                33.0 &       43.3 &     \valbest{61.3} \\
Ideology   &      33.3 &    64.8 &                    18.6 &                   23.7 &                    43.0 &                 47.6 &                  53.1 &               46.4 &            39.7 &                25.1 &              25.2 &                23.1 &                46.0 &                46.8 &       43.1 &     \valbest{60.0} \\
Impl. Hate &      16.7 &    62.5 &                     7.4 &                   14.4 &                     7.2 &                 \valbest{32.3} &                  29.6 &               32.0 &             7.1 &                 7.8 &               4.9 &                 9.2 &                18.4 &                19.2 &       16.3 &      3.7 \\
Misinfo    &      50.0 &    81.6 &                    33.3 &                   53.2 &                    64.8 &                  68.7 &                  69.6 &               \valbest{77.4} &            45.8 &                36.2 &              41.5 &                42.3 &                70.2 &                73.7 &       55.0 &     26.9 \\
Persuasion &      14.3 &    52.0 &                     3.6 &                   10.4 &                    37.5 &                 32.1 &                  45.7 &               43.5 &             3.6 &                 5.3 &               4.7 &                11.3 &                21.6 &                17.5 &       23.3 &     \valbest{56.4} \\
Sem. Chng. &      50.0 &    62.3 &                    33.5 &                   41.0 &                    \valbest{56.9} &                 52.0 &                  36.3 &               41.6 &            32.8 &                38.9 &              41.3 &                35.7 &                41.9 &                37.4 &       44.2 &     21.2 \\
Stance     &      33.3 &    36.1 &                    25.2 &                   36.6 &                    42.2 &                 43.2 &                  49.1 &               48.1 &            18.1 &                17.7 &              17.2 &                35.6 &                46.4 &                41.3 &       48.0 &     \valbest{76.0} \\
         \midrule
          \multicolumn{16}{c}{\textbf{Conversation Level Tasks}} \\
          \midrule
          Discourse  &      14.3 &     49.6 &                     4.2 &                   21.5 &                    33.6 &                 37.8 &                  \valbest{50.6} &               39.6 &             6.6 &                 9.6 &               4.3 &                11.4 &                35.1 &                36.4 &       35.4 &     16.7 \\
Empathy    &      33.3 &     71.6 &                    16.7 &                   16.7 &                    22.1 &                 21.2 &                  \valbest{35.9} &               34.7 &            24.5 &                17.6 &              27.6 &                16.8 &                16.9 &                17.4 &       22.6 &      6.4 \\
Persuasion &      50.0 &    33.3 &                     9.2 &                   11.0 &                    11.3 &                  8.4 &                  41.8 &               43.1 &             6.9 &                 6.7 &               6.7 &                33.3 &                33.3 &                \valbest{53.9} &       51.7 &     28.6 \\
Politeness &      33.3 &    75.8 &                    22.4 &                   42.4 &                    44.7 &                 57.2 &                  51.9 &               53.4 &            16.7 &                17.1 &              33.9 &                22.1 &                33.1 &                39.4 &       51.1 &     \valbest{59.7} \\
Power      &      49.5 &    72.7 &                    46.6 &                   48.0 &                    40.8 &                 55.6 &                  52.6 &               \valbest{56.9} &            43.1 &                39.8 &              37.5 &                36.9 &                39.2 &                51.9 &       56.5 &     42.0 \\
Toxicity   &      50.0 &    64.6 &                    43.8 &                   40.4 &                    42.5 &                 43.4 &                  34.0 &               48.2 &            41.4 &                34.2 &              33.4 &                34.8 &                41.8 &                46.9 &       31.2 &     \valbest{55.4} \\   
            
         \midrule
          \multicolumn{16}{c}{\textbf{Document Level Tasks}} \\
          \midrule
Event Arg. &      22.3 &     65.1 &                     -- &                   -- &                    -- &                  -- &                   -- &                -- &             -- &                 -- &               8.6 &                 8.6 &                21.6 &                22.9 &       22.3 &     \valbest{23.0} \\
Event Det. &  0.4 &   75.8 & 9.8 & 7.0 & 1.0 & 10.9 & 41.8 & 50.6 & 29.8 & 47.3 & 47.4 & 44.4 & 48.8 & \valbest{52.4} & 51.3 & 14.8\\
Ideology   &      33.3 &    85.1 &                    24.0 &                   19.2 &                    28.3 &                 29.0 &                  42.4 &               38.8 &            22.1 &                26.8 &              18.9 &                21.5 &                42.8 &                43.4 &       44.7 &     \valbest{51.5} \\
Tropes     &      36.9 &     - &                     1.7 &                    8.4 &                    13.7 &                 14.6 &                  19.0 &               28.6 &             7.7 &                12.8 &              16.7 &                15.2 &                16.3 &                26.6 &       36.9 &     \valbest{44.9} \\	
         
    \end{tabular}
    }
    \caption{\textbf{Zero-shot Classification Results} across our selected CSS benchmark tasks. All tasks are evaluated with \revision{macro F-1, which is averaged across 5 prompt permutations for zero-shot models. Supervised baseline results are averaged over 3 random seeds.} Best zero-shot models are in \valbest{green}. A dash indicates a model did not follow instructions.}
    \label{tab:clf_results}
\end{table*}

\begin{table}[]
    \centering
    \resizebox{\textwidth}{!}{%
    \begin{tabular}{l|lrrr}
    \toprule
Dataset & Best Model & F1 & $\kappa$ & Agreement \\[2pt] \midrule 
\multicolumn{5}{c}{\textbf{Utterance-Level}}\\\midrule
Dialect & flan-ul2 & 32.9 & 0.15 & \valbad{poor}\\ 
Emotion & flan-ul2 & \textbf{70.8} & 0.65 & \valbest{good}\\ 
Figurative & flan-ul2 & 62.3 & 0.52 & \valgood{moderate}\\  
Humor & gpt-4 & 61.3 & 0.23 & \valmid{fair}\\ 
Ideology & davinci-002 & 60.0 & 0.40 & \valgood{moderate}\\ 
Impl. Hate & flan-ul2 & 32.3 & 0.20 & \valmid{fair}\\ 
Misinfo & flan-ul2 & \textbf{77.4} & 0.55 & \valgood{moderate}\\ 
Persuasion & gpt-4 & 56.4 & 0.51 & \valgood{moderate}\\ 
Semantic Chng. & flan-t5-large & 56.9 & 0.14 & \valbad{poor}\\ 
Stance & {gpt-3.5-turbo} & \textbf{72.0} & 0.58 & \valgood{moderate}\\
    \end{tabular}
    \quad
    \begin{tabular}{l|lrrr}
    \toprule
        Dataset & Best Model & F1 & $\kappa$ & Agreement \\ \midrule 
        \multicolumn{5}{c}{\textbf{Convo-Level}}\\\midrule
        Discourse & flan-t5-xxl & 50.6 & 0.45 & \valgood{moderate}\\ 
        Empathy & flan-t5-xxl & 35.9 & 0.04 & \valbad{poor}\\ 
        Persuasion & davinci-003 & 53.9 & 0.14 & \valbad{poor}\\ 
        Politeness & flan-t5-xl & 59.2 & 0.38 & \valmid{fair}\\ 
        Power & gpt-4 & 59.7 & 0.26 & \valmid{fair}\\ 
        Toxicity & gpt-4 & 55.4 & 0.11 & \valbad{poor}\\\midrule
        \multicolumn{5}{c}{\textbf{Document-Level}}\\ \midrule
        Ideology & gpt-4 & 51.5 & 0.51 & \valgood{moderate}\\ 
        Event Det. & gpt-4 & 23.0 & n/a & -\\
        Tropes & gpt-4 & 44.9 & n/a & -\\
    \end{tabular}
    }
    \caption{\textit{(Acc.)} \textbf{Best model F1 scores.} \revision{F1 scores} above {70\%} are {bolded}. \textit{($\kappa$)} \textbf{Agreement scores between zero-shot model classification and human gold labels.} Out of ten utterance-level tasks, six have at least moderate \valgood{\tiny M} and only two have poor agreement \valbad{\tiny P}. Three (50\%) of the conversation tasks have at least fair agreement \valmid{\tiny F}.}
    \label{tab:kappa_classification}
\end{table}

Table~\ref{tab:clf_results} presents all zero-shot results for utterance, conversation, and document-level tasks. We use these results to answer Research Questions 1-3. The results suggest that \llm{s} are a \textbf{viable tool for augmenting human CSS annotation}. For classification tasks, results show that \textbf{larger, instruction-tuned open-source \llm{s} are preferable.}

\subsection{Viability (RQ1)}
\label{subsub:rq1}

\subsubsection{Zero-Shot Viability}
\label{subsub:zero_shot_viability}
\revision{To understand the viability of \llm{s} as CSS tools, we ask if zero-shot \llm{s} match or exceed the reliability of human annotation. If the overall performance is high and the expected agreement between humans and prompted models is as high as that between humans alone, then we might expect \llm{s} to viably augment the human annotation process. According to one paradigm, an \llm{} can serve as just one of many human and AI labelers, and gold labels would be decided by majority vote across these independent annotations. According to another complementary paradigm, \llm{} pseudo-labels can be used with a small set of gold-labels to compute unbiased estimators in downstream regressions following methods like Design-based Semi-supervised Learning} \citep[DSL;][]{egami2023using} and other .

\begin{figure}
    \centering
    \includegraphics[width=0.95\textwidth]{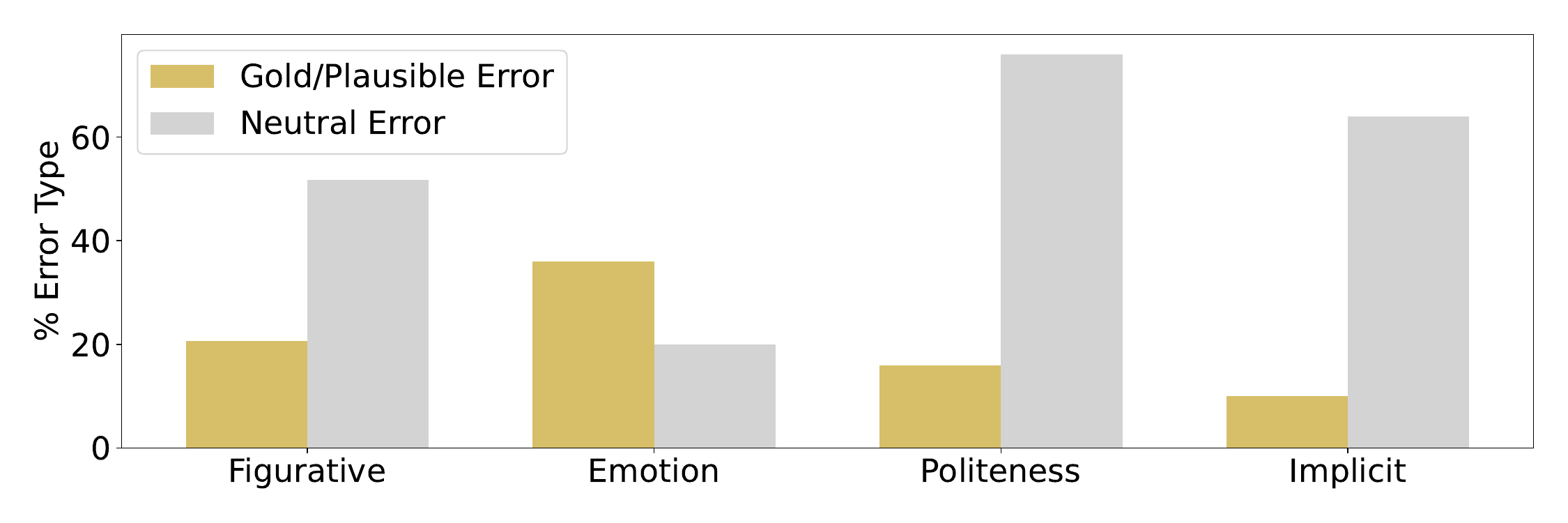}
    \caption{\textbf{Breakdown of Shared Error Types.} For a representative subset of classification tasks, we conduct an analysis of up to 50 shared errors across evaluated models. We focus specifically on the best performing model in a class (e.g. the best variant of FLAN models or the best OpenAI model). Plausible/gold errors occur when gold labels are incorrect or the model identifies a valid secondary label. Neutral errors occur when a model over-predicts a category in a respective task (\textit{metaphor} in Figurative; \textit{surprise} in Emotion; \textit{neutral} in Politeness, and \textit{stereotypical} in Implicit).}
    \label{fig:error_breakdown}
\end{figure}

\revision{Table~\ref{tab:clf_results} shows the best zero-shot models achieve as high as 77.4 F1 on misinformation detection. On this task, the best-performing FLAN-UL2 model achieves an agreement of $\kappa=0.55$ with gold labels (see Table~\ref{tab:kappa_classification}), which is higher than the $\kappa=0.51$ inter-human agreement reported by \citet{gabriel2022misinfo} in their original paper. In fact, for 8/17 tasks in Table~\ref{tab:kappa_classification}, or 47\% of classification tasks, models achieve moderate to good agreement scores ranging from $\kappa=0.40$ to $0.65$. These tasks also correspond to the highest absolute performances on \textit{Stance} (76.0 F1) \textit{Emotion} (70.8 F1), \textit{Figurative Language} (62.3 F1) and utterance-level \textit{Ideology Classification} (60.0 F1). \textbf{In these cases of high viability, we recommend that CSS researchers consider the DSL and augmented-annotator paradigms}. Such strong performances are on tasks that either have objective ground truth (fact checking for misinformation) or have labels with explicit colloquial definitions in the pretraining data (emotional categories like \textit{anger} are part of everyday vernacular; political stances are well-documented and explicit in online forums). Our qualitative error analysis in \S\ref{subsub:error_analysis} will show that here, models are less likely to default to neutral categories, and errors are more likely to come from annotation mistakes in the gold dataset according to the author's own manual error analysis (see lower neutral and higher gold error in Figure~\ref{fig:error_breakdown}).}

Are zero-shot models ready to label text out-of-the-box? Zero-shot results rarely exceed the carefully-tuned supervised RoBERTa baselines in Table~\ref{tab:clf_results}. However, the best observed performances here match that of classifiers used in published studies on stance detection \citep[67.8 F1;][]{zarrella2016mitre}, COVID-19 vaccination opinions \citep{cotfas2021longest}, political opinions \citep{siddiqua2019tweet}, and debates \citep{lai2020multilingual}. In such scenarios, \textbf{zero-shot models could offer a data-efficient alternative to fine-tuned models} by removing the need for expensive training sets. Humans could focus all of their efforts on validating \llm{} outputs and tuning prompts (\S\ref{subsec:prompt_engineering}) rather than coding unstructured text. However, we encourage practitioners to proceed cautiously, especially in sensitive domains, and we recommend human-in-the-loop methods to mitigate bias and risk. See \S\ref{subsub:bias_fairness} and \S\ref{subsub:ethical_considerations} for further discussion.

\revision{It is important to consider that \llm{} performance could be unusably poor for some CSS tasks. For a non-negligible subset of tasks we considered, \llm{s} have poor agreement (5/17 = 29.5\%), and here social scientists might not consider zero-shot annotation augmentation via \llm{s}. On these poor agreement tasks, \llm{} absolute performance is not significantly better than random guessing: see 56.9 F1 vs 50 Random on \textit{Semantic Change}; 35.9 F1 vs 33.3 Random on \textit{Empathy}; 53.9 F1 vs 50 Random on 
conversation-level \textit{Persuasion}; and 55.4 F1 vs 50 Random on \textit{Toxicity}. Some of these low-performance tasks like \textit{Event Argument Extraction} are structurally complex and may require additional engineering efforts. Others like \textit{Empathy} and \textit{Tropes} have challenging and subjective expert taxonomies whose semantics differ from definitions learned in model pretraining. This is confirmed by our error analysis in Figure~\ref{fig:error_breakdown} where GPT3.5 often defaults to the neutral, more colloquially recognizable label \textit{stereotype} (64\% of errors) rather than use a more taxonomy-specific label like \textit{white grievance}. In \S\ref{subsub:few_shot_viability}, we test if few-shot prompting can reduce misalignments between model and ground-truth definitions.}

\subsubsection{Zero-Shot Error Analysis}
\label{subsub:error_analysis}
\revision{For a representative subset of classification tasks, we conduct an analysis of shared errors across evaluated models. We focus specifically on the best performing model in a class (e.g. the best variant of FLAN models or the best OpenAI model). Finally, in Figure~\ref{fig:error_breakdown} we break down the error types for \texttt{\small gpt-3.5-turbo}. }

\paragraph{Figurative Language}
\revision{We sample all 29 cases in which every model was incorrect. In just under half of these cases (14/29), all models agreed on an incorrect answer, which we call a \textit{unanimous error}. Out of fourteen unanimous errors, the models were at least partially correct four times, which we call a \textit{plausible/gold error} (see Figure~\ref{fig:error_breakdown}). There was one mistaken gold label and three cases of correctly-labeled similes nested inside the predicted sarcasm. Of the remaining ten unanimous errors, three were idioms mistaken as metaphors, and seven were similes classified with the more general metaphor label. For humans, this is a common error, but for models, this is surprising, since similes should have easy keyword signals ``as'' and ``like.'' The baseline method was likely able to exploit these signals to achieve a higher accuracy.

In 5 errors, all models disagreed and missed the intended sarcasm label.
In another 5 error cases, only UL2 and \texttt{text-davinci-003} agreed on the correct label, but the dataset was mislabeled, with four idioms marked wrongly as metaphors and one simile marked as an idiom. In the remaining 5 errors, ChatGPT showed a preference for the most generic label and predicted metaphor.}

\paragraph{Emotion Recognition}
\revision{We sample 50 cases where all models differed from the gold labels. Unlike Figurative Language, a minority of examples had the same mismatch across models (9/50). However, a closer analysis of individual errors yields a surprising result: at least 18/50 examples \textit{across all evaluated models} were judged as gold mislabels. Additionally, for FLAN-UL2 and ChatGPT, 17/50 and 15/50 predictions respectively could be considered as valid---even if they differed from the gold label.\footnote{For example, ``\emph{i feel that the sweet team really accomplished that}'' can be considered both \textit{love - gold} or \textit{joy - predicted}} 

Moving to true negatives, we observe that DV2 makes the most errors (28/50) that cannot be categorized as a gold mislabel, while UL2 (17/20) and ChatGPT (19/20) make significantly fewer. The distribution of errors differ across each model type: ChatGPT, for example, over-labels with \textit{surprise}: especially instances with a true gold label of \textit{Joy} (8) or \textit{Love} (5). On the other hand, UL2 mislabels \textit{Love} as \textit{Joy} frequently (9); and fear as \textit{Sadness} (4) or \textit{Surprise} (4). Finally, \texttt{davinci} mislabels Sadness most frequently as Joy (9) or anger/love (3 each).}

\paragraph{Politeness Prediction}
\revision{We first visualize the per-category accuracy of the different best-performing models (FLAN-T5-XL, \texttt{Text-davinci-002}, and ChatGPT). We observe that: (1) The XL model tends to predict more polite labels. It is more accurate in terms of the utterances that were polite and neutral with 70.4\% and 62.0\% accuracy. Most errors come from impolite cases (with a 45.2\% accuracy). (2) \texttt{davinci-002} performs best in judging neutral utterances. \texttt{davinci-002} is the most accurate for neutral utterances (82.9\% accuracy) while making significantly more errors for polite and impolite utterances (43.9\% and 40.9\% accuracy respectively). (3) ChatGPT performs worst in finding impolite utterances while making more neutral predictions, with only 9.0\% accuracy for the impolite category, whereas it achieves 75.9\% and 66.8\% for neutral and polite cases. 

We then consider the 81/498 cases where the three models are all making errors. We find that the three models make the same errors in most cases (54/81) and \texttt{davinci-002} models make errors more similar to ChatGPT (17/81 cases). Among these common error cases, we observe that 79/81 cases are related to the 1st and 2nd person mention strategy \citep{danescu-niculescu-mizil-etal-2013-computational} and all of them are direct or indirect questions, while 38/81 are related to counterfactual modal and indicative modal \citep{danescu-niculescu-mizil-etal-2013-computational}. This indicates that all three models struggle with direct or indirect questions with 1st and 2nd person mentions.}

\paragraph{Implicit Hate Classification}
\revision{We first consider the confusion matrix and find that OpenAI models are particularly oversensitive to the ``stereotypical'' class (71\% and and 65\% false-positive rates from \texttt{davinci-003} and ChatGPT respectively). Our error analysis of 50 samples shows that models fail to apply the definition: stereotypical text must associate the target with particular characteristics. Instead, models are more likely to mark as stereotype any text that contains an identity term (86\% of false-positives contain identity terms). All models also fail to recognize strong phrasal signals, like ``rip'' or ``kill white people'' for the \textit{white grievance} (all 3/50 cases are errors), or violent terms associated with threats. More subtle false-negatives require sociopolitical knowledge (2/50) or understanding of humor (6/50). Other errors are examples where the model identified a valid secondary hate category (5/50).}

\subsubsection{Few-Shot Viability}
\label{subsub:few_shot_viability}
\revision{Zero-shot models may not be naturally aligned with the non-standard or technical meanings of certain key terms in the social sciences. To address this issue, we consider the viability of open-source FLAN models for few-shot classification. Specifically, we try 3-shot and 5-shot experiments with no further prompt engineering. Table~\ref{tab:fewshot_results} shows that any improvements from these methods are spotty and inconsistent. For some challenging tasks like Empathy and Persuasion that have subjective definitions or non-standard taxonomies, few-shot learning can improve performance in 2 and 5 out of 6 model sizes respectively. However, these gains are small and not widespread among other tasks. We conclude that \textbf{additional engineering efforts may be needed to achieve significant gains on CSS tasks via few-shot learning.}}

\begin{table*}[]
    \centering
    \resizebox{\textwidth}{!}{%
    \def\arraystretch{1.15}
    \setlength{\tabcolsep}{4pt}
    \begin{tabular}{l|rrr|rrr|rrr|rrr|rrr|rrr}
         \toprule 
          \textbf{Model}
          & \multicolumn{3}{c}{FLAN Small} &  \multicolumn{3}{c}{FLAN Base} &
          \multicolumn{3}{c}{FLAN Large} &
          \multicolumn{3}{c}{FLAN XL} &
          \multicolumn{3}{c}{FLAN XXL} &
          \multicolumn{3}{c}{FLAN UL2} \\
         \cmidrule(lr){2-4} \cmidrule(lr){5-7} \cmidrule(lr){8-10} \cmidrule(lr){11-13} \cmidrule(lr){14-16} \cmidrule(lr){17-19}
         \textbf{Shot} & 0\phantom{0} & 3\phantom{0} & 5\phantom{0} & 0\phantom{0} & 3\phantom{0} & 5\phantom{0}& 0\phantom{0} & 3\phantom{0} & 5\phantom{0}& 0\phantom{0} & 3\phantom{0} & 5\phantom{0}& 0\phantom{0} & 3\phantom{0} & 5\phantom{0}& 0\phantom{0} & 3\phantom{0} & 5\phantom{0}\\
 \midrule
         Dialect    &                     0.2 &                     0.0 &                     \textbf{0.4} &                    \textbf{4.5} &                    0.0 &                    1.4 &                    \textbf{23.4} &                     0.7 &                    14.1 &                 \textbf{24.8} &                  8.0 &                 20.5 &                  \textbf{30.3} &                   0.2 &                  29.9 &               \textbf{32.9} &               12.6 &               27.5 \\
Emotion    &                    \textbf{19.8} &                    10.6 &                    10.1 &                   \textbf{63.8} &                   42.7 &                   42.0 &                    \textbf{69.7} &                    67.6 &                    67.4 &                 \textbf{65.7} &                 62.1 &                 62.5 &                  \textbf{66.2} &                  61.8 &                  57.4 &               \textbf{70.8} &               70.0 &               69.8 \\
Figurative &                    \textbf{16.6} &                    10.0 &                     9.2 &                   23.2 &                   \textbf{29.1} &                   27.3 &                    18.0 &                    \textbf{21.8} &                    19.6 &                 \textbf{32.2} &                 27.9 &                 28.5 &                  53.2 &                  52.6 &                  \textbf{66.2} &               \textbf{62.3} &               52.7 &               62.0 \\
Humor      &                    51.8 &                    52.8 &                    \textbf{53.1} &                   \textbf{37.1} &                   35.1 &                   34.7 &                    \textbf{54.9} &                    54.0 &                    53.8 &                 56.9 &                 \textbf{57.0} &                 56.7 &                  29.9 &                  34.8 &                  \textbf{35.3} &               \textbf{56.8} &               55.5 &               54.1 \\
Ideology   &                    18.6 &                    16.7 &                    \textbf{24.0} &                   \textbf{23.7} &                   22.6 &                   38.3 &                    43.0 &                    \textbf{47.3} &                    45.5 &                 47.6 &                 \textbf{48.8} &                 50.4 &                  53.1 &                  52.9 &                  \textbf{57.7} &               46.4 &               36.9 &               \textbf{51.5} \\
Impl. Hate &                     \textbf{7.4} &                     6.8 &                     6.2 &                   14.4 &                   \textbf{21.1} &                    7.4 &                     7.2 &                     \textbf{9.3} &                     4.7 &                 32.3 &                 28.5 &                 \textbf{34.6} &                  29.6 &                  31.6 &                  \textbf{35.1} &               \textbf{32.0} &               29.5 &               25.9 \\
Misinfo    &                    \textbf{33.3} &                    33.3 &                    33.3 &                   53.2 &                   45.3 &                   \textbf{59.7} &                    \textbf{64.8} &                    64.8 &                    64.2 &                 68.7 &                 67.2 &                 \textbf{69.7} &                  69.6 &                  \textbf{74.9} &                  74.4 &               \textbf{77.4} &               53.7 &               76.4 \\
Persuasion &                     \textbf{3.6} &                     3.6 &                     3.6 &                   10.4 &                   \textbf{10.8} &                    7.3 &                    37.5 &                    \textbf{39.0} &                    37.7 &                 32.1 &                 \textbf{44.3} &                 41.8 &                  45.7 &                  44.6 &                  \textbf{48.6} &               \textbf{43.5} &               42.2 &               40.1 \\
Sem. Chng. &                    33.5 &                    33.3 &                    \textbf{34.0} &                   41.0 &                   35.7 &                   \textbf{41.7} &                    56.9 &                    48.8 &                    \textbf{60.4} &                 \textbf{52.0} &                 40.8 &                 35.6 &                  \textbf{36.3} &                  34.0 &                  33.3 &               41.6 &               \textbf{62.5} &               34.6 \\
Stance     &                    25.2 &                    16.7 &                    \textbf{29.6} &                   \textbf{36.6} &                   18.1 &                   36.6 &                    \textbf{42.2} &                    41.8 &                    39.8 &                 43.2 &                 \textbf{52.1} &                 46.2 &                  \textbf{49.1} &                  46.0 &                  48.7 &               48.1 &               \textbf{55.6} &               54.7 \\
Discourse  &                     4.2 &                     4.0 &                     \textbf{7.5} &                   \textbf{21.5} &                   18.1 &                   20.7 &                    33.6 &                     3.6 &                    \textbf{34.6} &                 37.8 &                  3.6 &                 \textbf{38.0} &                  \textbf{50.6} &                   3.6 &                  43.4 &               \textbf{39.6} &                3.6 &               39.1 \\
Empathy    &                    \textbf{16.7} &                    16.7 &                    16.7 &                   \textbf{16.7} &                   16.7 &                   16.7 &                    \textbf{22.1} &                    16.7 &                    17.1 &                 21.2 &                 \textbf{30.4} &                 22.8 &                  \textbf{35.9} &                  29.8 &                  28.2 &               34.7 &               \textbf{41.5} &               39.6 \\
Persuasion &                     9.2 &                    \textbf{55.9} &                    45.0 &                   11.0 &                   \textbf{55.0} &                   48.7 &                    11.3 &                    \textbf{54.6} &                    51.7 &                  8.4 &                 42.8 &                 \textbf{43.8} &                  \textbf{41.8} &                  38.8 &                  35.2 &               43.1 &               \textbf{44.9} &               46.1 \\
Politeness &                    \textbf{22.4} &                    16.7 &                    20.1 &                   \textbf{42.4} &                   23.9 &                   35.4 &                    44.7 &                    44.5 &                    \textbf{51.9} &                 \textbf{57.2} &                 27.7 &                 50.4 &                  \textbf{51.9} &                  44.2 &                  50.3 &               53.4 &               43.6 &               \textbf{53.9} \\
Power      &                    \textbf{46.6} &                    44.5 &                    33.3 &                   \textbf{48.0} &                   39.8 &                   41.4 &                    40.8 &                    \textbf{45.5} &                    43.5 &                 55.6 &                 58.9 &                 \textbf{60.2} &                  52.6 &                  52.0 &                  \textbf{62.6} &               56.9 &               57.2 &               \textbf{57.5} \\
Toxicity   &                    43.8 &                    \textbf{46.7} &                    33.3 &                   40.4 &                   34.7 &                   \textbf{54.4} &                    \textbf{42.5} &                    34.7 &                    36.7 &                 43.4 &                 38.7 &                 \textbf{49.2} &                  34.0 &                  33.3 &                  \textbf{35.1} &               48.2 &               44.7 &               \textbf{52.5} \\
Ideology   &                    \textbf{24.0} &                    16.7 &                    19.2 &                   19.2 &                   16.6 &                   \textbf{21.3} &                    \textbf{28.3} &                    17.0 &                    17.9 &                 29.0 &                 \textbf{31.7} &                 27.0 &                  42.4 &                  \textbf{48.5} &                  47.9 &               38.8 &               \textbf{38.9} &               39.7 \\
Tropes     &                     1.7 &                     \textbf{5.1} &                     3.4 &                    \textbf{8.4} &                    5.1 &                    3.4 &                    \textbf{13.7} &                    10.0 &                    11.6 &                 \textbf{14.6} &                  8.4 &                 10.0 &                  \textbf{19.0} &                   8.4 &                   6.8 &               \textbf{28.6} &               27.3 &               24.6
    \end{tabular}
    }
    \caption{\revision{\textbf{Few-shot Classification} does not uniformly improve performance across our selected CSS benchmark tasks. All tasks are evaluated with macro F-1.}}
    \label{tab:fewshot_results}
\end{table*}

\subsection{Model-Selection (RQ2)}
\label{subsub:rq2}

CSS researchers should understand how their choice of model can decide the reliability of zero-shot methods. Our results show that, for structured parsing tasks like event extraction, it is best to use a code-instructed model like OpenAI's \texttt{gpt-3.5-turbo}, while for most classification tasks, {open-source \llm{s} like FLAN-UL2 are best}.

\paragraph{Model Size} 
\llm{s} generally follow scaling laws~\citep{kaplan2020scaling, hoffmann2022training} where performance increases with the size of the model and training data. We investigate scaling laws in the two families of instruction-tuned \llm{s}: FLAN and OpenAI. Results show larger open-sourced models are preferable.

\begin{figure}[]
    \centering
    \includegraphics[trim={0.75cm 0.8cm 0cm 0.8cm},clip,width=0.9\textwidth]{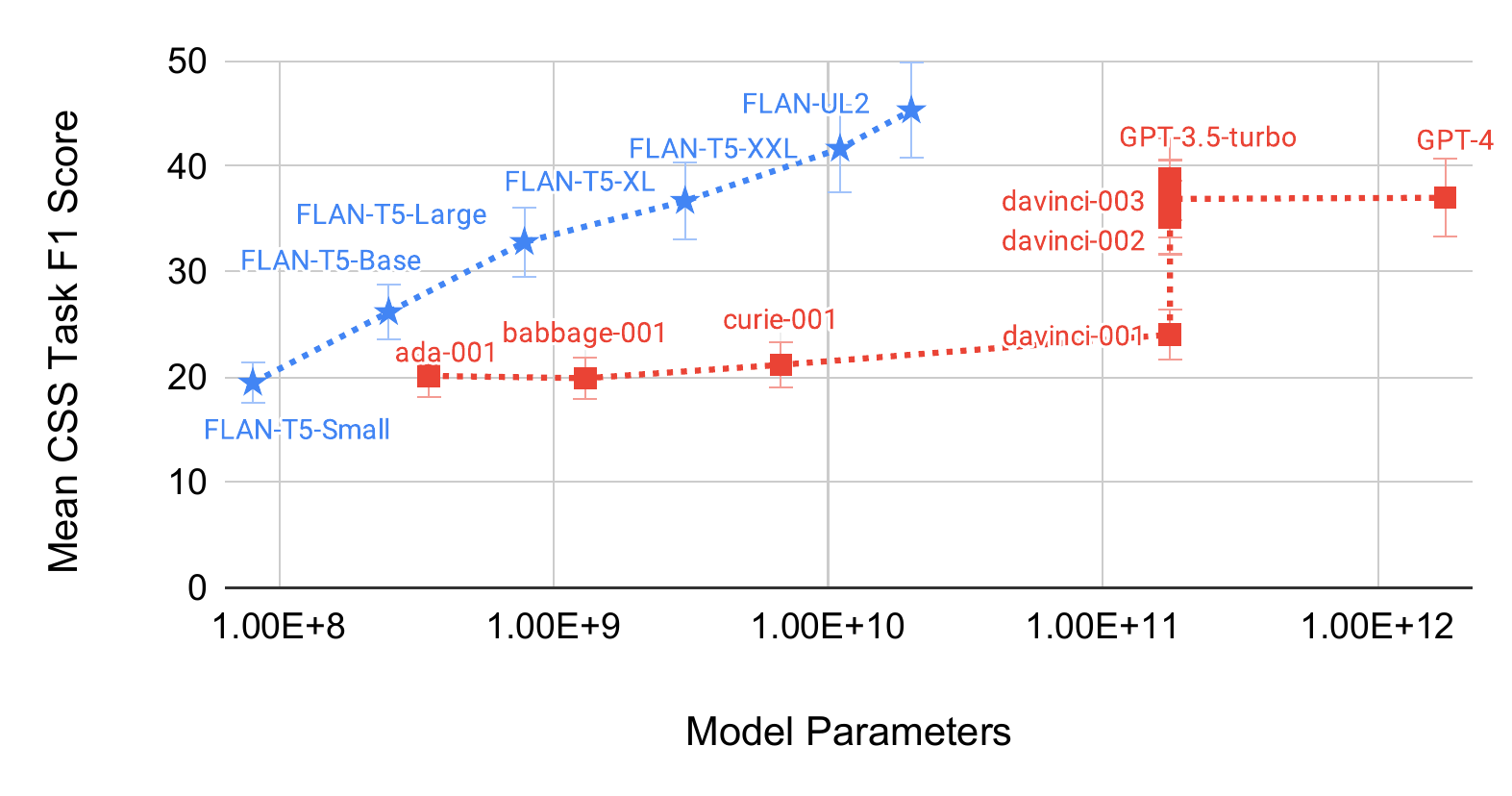}
    \caption{\textbf{Effects of Scaling} on the mean zero-shot performance on our CSS benchmark tasks. FLAN models and \texttt{davinci-001/002} are instruction fine-tuned. \texttt{davinci-003} and \texttt{gpt-3.5-turbo} are instruction fine-tuned and refined with Reinforcement Learning from Human Feedback. GPT Parameter counts reported based on approximates.}
    \label{fig:model_scaling}
\end{figure}

\textsl{FLAN's CSS performance roughly matches \citeauthor{kaplan2020scaling}'s predicted power-law effects from pure model size}. Figure~\ref{fig:model_scaling} shows FLAN classification performances scaling nearly logarithmically with the parameter count. With each order of magnitude size increase, the median average task improvement in FLAN models is 5.0 absolute percentage points. All FLAN-T5 models use the same stable corpus, pretraining objective, and architecture, which gives us a controlled environment to observe stable scaling laws.

\textsl{OpenAI's GPT-3 \texttt{001} models, on the other hand, do not monotonically benefit from scaling.}\footnote{This analysis relies on estimates which combine \href{https://blog.eleuther.ai/gpt3-model-sizes/}{community estimates}, the \href{https://platform.openai.com/docs/model-index-for-researchers}{OpenAI research documentation}, and the assumption that all models named or "improved" from \texttt{davinci} have the same parameter counts. These estimates may be incorrect, as hypothesized by other \href{https://orenleung.com/is-chatgpt-175-billion-parameters-technical-analysis}{community estimates}. This is a limitation of research on these models as exact model size and training data are a trade secret of OpenAI.} There is minimal performance improvement from \texttt{ada} to \texttt{davinci-001}, despite the three orders of magnitude increase in size. Similarly, we observe little benefit from the trillion-parameter GPT-4 over the hundred-billion-parameter GPT-3.5-turbo. Instead, the greatest performance improvements come from variations in \textit{pretraining, fine-tuning}, and \textit{reinforcement learning}.

\paragraph{Pretraining \& Instruction Fine-tuning}  Besides scale, two key factors play a major role in model performance: \emph{pretraining data} and \emph{instruction fine-tuning}. Pretraining data is the raw text upon which an \llm{} learns to model the general generative process of language. Instruction fine-tuning refines the raw \llm{} to perform specific tasks based on human-written instructions.

\textsl{OpenAI's \texttt{davinci} models significantly benefit from pretraining and instruction fine-tuning tricks.} For classification tasks (Table~\ref{tab:clf_results}), we see an outsized increase in CSS performance ($\uparrow$ absolute 11 pct.~pts.) moving from \texttt{davinci-001} to \texttt{davinci-002}, larger than any performance increase from scale alone. Both \texttt{davinci-001} and \texttt{davinci-002} use the same supervised instruction fine-tuning strategy, but \texttt{davinci-002} is based on OpenAI's base-code model, which had access to a larger set of instruction fine-tuning data. Most importantly, \texttt{davinci-002} was pre-trained on both text and code. This difference benefits structured, JSON-formatted tasks like Event Argument Extraction. While \texttt{davinci-001} often fails to generate JSON, \texttt{davinci-002} succeeds with markedly improved performance (+13.0 F1).

\paragraph{Learning From Human Feedback} We see that \textit{RLHF can improve \llm{} performance on CSS classification tasks.} RLHF has been lauded as the major catalyst behind the success of instruction-following models~\citep{ouyang2022training}, and here we see \texttt{text-davinci-003} and \texttt{gpt-3.5-turbo} (with RLHF) improves the average F1 of  \texttt{text-davinci-002} (without RLHF) by 3.5 absolute points. 

\subsection{Domain-Utility (RQ3)}
\label{subsub:rq3}
The survey and taxonomy of social science need in \S\ref{sec:core_methods_css} allows us to understand whether the utility of \llm{s} is limited to certain domains or certain data types. To do so, we partition all classification results from Table~\ref{tab:clf_results} into bins corresponding to the academic field most impacted by the task.\footnote{This partitioning follows Figure~\ref{fig:crown_jewel}, with stance and ideology detection in the \textit{political science} bin and {dialect feature classification} under \textit{linguistics}, for example.} Although we recognize the multi-disciplinary utility of \textit{all} tasks, this type of 1:1 organization is appropriate for understanding the academic scope of our results. We acknowledge that the partitioning and selection of the dataset influence the performance distributions that we observe. We urge readers to interpret the results with caution and focus on broader conclusions rather than the fine numerical details of these distributions.

\begin{figure}[]
    \centering
        \includegraphics[width=0.57\textwidth]{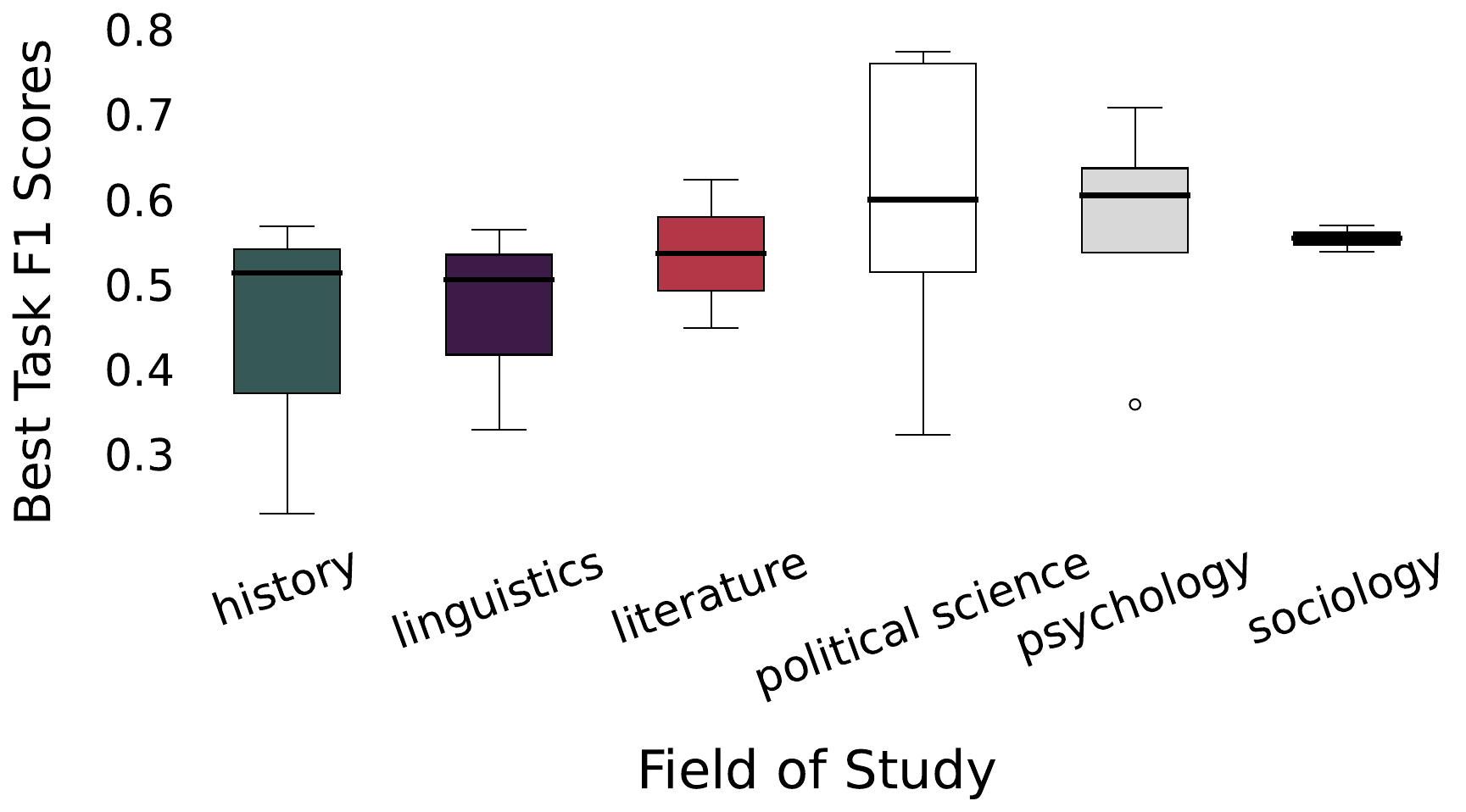}
    \rulesep
        \includegraphics[width=0.38\textwidth]{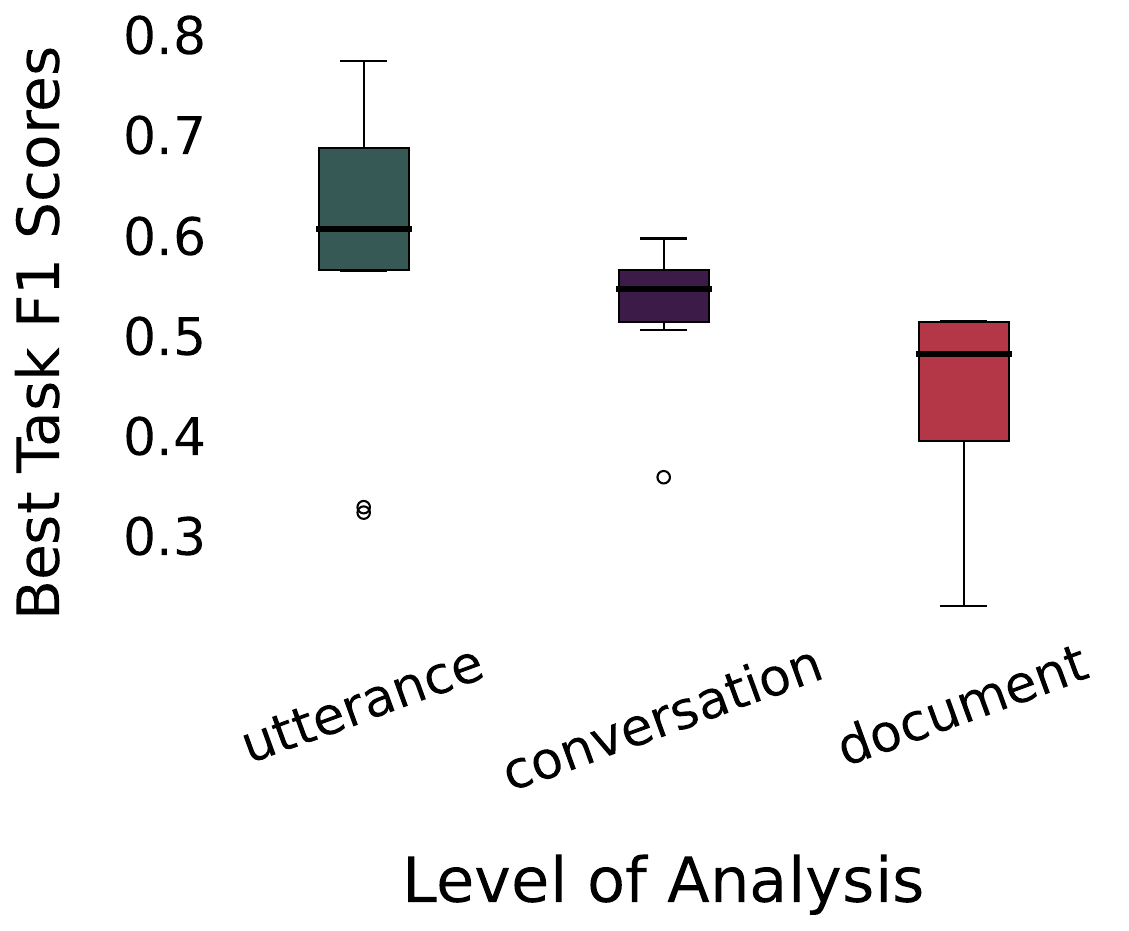}
    \caption{(\textit{Left}) \textbf{Task Performance By Field of Study}. Significant overlap in the distributions suggests that neither high nor low performance is exclusive to any particular discipline. \textit{Caution:} The distributions depend on the particular choices of this study, which datasets to select and how to partition them. \newline (\textit{Right}) \textbf{Task Performance By Level of Analysis}. Document-level tasks are challenging for their input length and complexity, and this is reflected in their F1 scores all near or below 50\%. Utterance and conversation-level task performance varies also with the complexity of the task.}
    \label{fig:performance_by_field_level}
\end{figure}

The box plot in Figure~\ref{fig:performance_by_field_level} shows that field-specific performances significantly overlap. Thus overall, \textbf{we do not observe a strong bias against or proclivity for a particular field of study.}
In political science, we see the highest overall performance on misinformation detection (F1=77.4) and much lower performance on implicit hate detection (F1=32.3). For psychology, we observe high performance on emotion detection (F1=70.8) and low performance on empathy detection (F1=35.9).  Peformances span the full range of disciplines. This suggests that performance is not tied to academic discipline.

In terms of data type, Figure~\ref{fig:performance_by_field_level} suggests that \textbf{performance may be more closely determined by the complexity of the input}. In particular, documents encode complex sequences of ideas or temporal
events, and overall, the two lowest task performances are on the document-level tasks: character trope classification and event argument extraction. All other document-level accuracies are at or below 50\%. 
The most challenging utterance and conversation-level tasks are also a function of their label space complexity. Implicit hate (F1=32.3), empathy (F1=35.9), and dialect feature (F1=32.9) annotations are expert-labeled on a subtle theoretical taxonomy.

\section{Generation Results}
\label{subsec:gen_results}
In this section, we answer \textit{RQ4: Are prompted \llm{s} useful for generatively implementing theories and explaining social scientific constructs with text?} Will generative models replace or augment human analysis? To answer this question, we rely on the human evaluation setup described in \S\ref{subsub:human_eval_method}. Note that FLAN models are excluded from all evaluation tables because FLAN models failed to follow instructions by manual inspection. Instead, we evaluate across the OpenAI suite.

\paragraph{Human Scoring Evaluation}
\revision{According to the domain experts in Table~\ref{tab:generation_finegrained_results}, leading generative models can produce text of a quality that matches or exceeds that of human gold references. For Aspect-Based Emotion Summarization, Misinformation Explanation, and Social Bias Inference, \texttt{gpt-3.5-turbo} produce, on average, more domain-faithful, coherent, and fluent text than both gold references and the fine-tuned baseline. For Positive Reframing and Figurative Language explanation, \texttt{text-davinci-003} match the gold-standard levels of faithfulness, relevance, coherence, and fluency, again outperforming the fine-tuned baseline. For generation performance, scale matters, as smaller models fail to produce explanations and summaries that are faithful to the task specifications, especially in \textsc{CovidET}, FLUTE, and SBIC, where \texttt{text-ada-001} earns faithfulness scores of less than 2 out of 5 on average. With scale, however, \textbf{\llm{s} are capable of generating useful, relevant, coherent, and fluent explanations and summaries of underlying social science constructs}.}

\begin{table}[]
    \centering
    \resizebox{\textwidth}{!}{%
    \begin{tabular}{l|cccc}
    \toprule
    \multicolumn{5}{c}{Aspect-Based Summarization (\textsc{CovidET})} \\ \midrule
    Model & Faithful & Relevant & Coherent & Fluent \\ \midrule 
Baseline         &        2.1\phantom{$^-$} &        2.3\phantom{$^-$} &           2.1$^-$ &           2.6$^-$ \\
\texttt{ada-001}     &           1.8$^-$ &           1.8$^-$ &        2.4\phantom{$^-$} &        3.6\phantom{$^-$} \\
\texttt{babbage-001} &           2.0$^-$ &        2.0\phantom{$^-$} &        2.3\phantom{$^-$} &        3.7\phantom{$^-$} \\
\texttt{curie-001}   &        2.3\phantom{$^-$} &        2.3\phantom{$^-$} &        2.6\phantom{$^-$} &        3.8\phantom{$^-$} \\
\texttt{davinci-001} &        2.3\phantom{$^-$} &        2.4\phantom{$^-$} &        2.5\phantom{$^-$} &        3.9\phantom{$^-$} \\
\texttt{davinci-002} &        2.4\phantom{$^-$} &        2.5\phantom{$^-$} &        3.2\phantom{$^-$} &        4.0\phantom{$^-$} \\
\texttt{davinci-003} &        2.9\phantom{$^-$} &        2.8\phantom{$^-$} &        3.0\phantom{$^-$} &  \textbf{4.1$^+$} \\
\texttt{GPT 3.5}          &  \valbest{\textbf{3.9$^+$}} &  \valbest{\textbf{3.5$^+$}} &  \valbest{\textbf{3.8$^+$}} &  \valbest{\textbf{4.5$^+$}} \\
\texttt{GPT 4}            &  \valgood{\textbf{3.7$^+$}} &  \valgood{\textbf{3.3$^+$}} &  \valgood{\textbf{3.8$^+$}} &  \valgood{\textbf{4.4$^+$}} \\
Human            &        2.8\phantom{$^-$} &        2.6\phantom{$^-$} &        2.8\phantom{$^-$} &        3.8\phantom{$^-$} \\ \midrule
    \multicolumn{5}{c}{Figurative Language Explanation (FLUTE)} \\ \midrule
    Model & Faithful & Relevant & Coherent & Fluent \\ \midrule 
Baseline         &      1.4$^-$ &     1.7$^-$ &     1.4$^-$ &        4.2\phantom{$^-$} \\
\texttt{ada-001}     &      1.4$^-$ &     1.5$^-$ &     1.5$^-$ &        3.9\phantom{$^-$} \\
\texttt{babbage-001} &      1.4$^-$ &     1.9$^-$ &     1.5$^-$ &           3.9$^-$ \\
\texttt{curie-001}   &      1.5$^-$ &     2.3$^-$ &     1.7$^-$ &        4.1\phantom{$^-$} \\
\texttt{davinci-001} &      1.2$^-$ &     1.9$^-$ &     1.5$^-$ &        4.1\phantom{$^-$} \\
\texttt{davinci-002} &   \valgood{2.5\phantom{$^-$}} &  3.4\phantom{$^-$} &  2.5\phantom{$^-$} &        4.1\phantom{$^-$} \\
\texttt{davinci-003} &   \valbest{3.0\phantom{$^-$}} &  \valbest{4.0\phantom{$^-$}} &  \valbest{3.1\phantom{$^-$}} &  \valbest{\textbf{4.1$^+$}} \\
\texttt{GPT 3.5}          &      2.1$^-$ &  \valgood{3.6\phantom{$^-$}} &  \valgood{2.5\phantom{$^-$}} &        \valgood{4.1\phantom{$^-$}} \\
\texttt{GPT 4}            &      2.1$^-$ &  3.3\phantom{$^-$} &  2.4\phantom{$^-$} &        4.0\phantom{$^-$} \\
Human            &   2.8\phantom{$^-$} &  4.0\phantom{$^-$} &  2.6\phantom{$^-$} &        4.2\phantom{$^-$} \\ \midrule
        \multicolumn{5}{c}{Positive Reframing} \\ \midrule
    Model & Faithful & Relevant & Coherent & Fluent \\ \midrule 
       Baseline         &   4.1\phantom{$^-$} &        4.2\phantom{$^-$} &  3.9\phantom{$^-$} &        4.4\phantom{$^-$} \\
\texttt{ada-001}     &      1.8$^-$ &           1.4$^-$ &     1.8$^-$ &           1.6$^-$ \\
\texttt{babbage-001} &   3.8\phantom{$^-$} &           2.5$^-$ &  3.8\phantom{$^-$} &        3.7\phantom{$^-$} \\
\texttt{curie-001}   &   4.1\phantom{$^-$} &           3.7$^-$ &  4.1\phantom{$^-$} &        3.9\phantom{$^-$} \\
\texttt{davinci-001} &      3.5$^-$ &        4.0\phantom{$^-$} &     3.3$^-$ &        4.1\phantom{$^-$} \\
\texttt{davinci-002} &   4.0\phantom{$^-$} &           3.9$^-$ &  4.0\phantom{$^-$} &        4.2\phantom{$^-$} \\
\texttt{davinci-003} &   \valbest{4.4\phantom{$^-$}} &  \valbest{\textbf{4.5$^+$}} &  \valbest{4.2\phantom{$^-$}} &  \valbest{\textbf{4.6$^+$}} \\
\texttt{GPT 3.5}          &   \valgood{4.3\phantom{$^-$}} &        \valgood{4.3\phantom{$^-$}} &  \valgood{4.2\phantom{$^-$}} &        \valgood{4.4\phantom{$^-$}} \\
\texttt{GPT 4}            &   4.1\phantom{$^-$} &        4.3\phantom{$^-$} &  4.1\phantom{$^-$} &        4.2\phantom{$^-$} \\
Human            &   4.2\phantom{$^-$} &        4.2\phantom{$^-$} &  4.1\phantom{$^-$} &        4.2\phantom{$^-$} \\
    \end{tabular}
    \quad
    \begin{tabular}{l|cccc}
    \toprule
    \multicolumn{5}{c}{Implied Misinformation Explanation (MRF)} \\ \midrule
    Model & Faithful & Relevant & Coherent & Fluent \\ \midrule 
Baseline         &        3.4\phantom{$^-$} &        3.5\phantom{$^-$} &        3.7\phantom{$^-$} &        4.2\phantom{$^-$} \\
\texttt{ada-001}     &           1.1$^-$ &           1.1$^-$ &           2.0$^-$ &        4.5\phantom{$^-$} \\
\texttt{babbage-001} &           1.6$^-$ &           1.7$^-$ &           2.5$^-$ &        4.3\phantom{$^-$} \\
\texttt{curie-001}   &           2.6$^-$ &           2.7$^-$ &           3.1$^-$ &        4.4\phantom{$^-$} \\
\texttt{davinci-001} &           1.7$^-$ &           1.7$^-$ &           2.5$^-$ &        4.5\phantom{$^-$} \\
\texttt{davinci-002} &  \valbest{\textbf{3.9$^+$}} &  \valbest{\textbf{4.1$^+$}} &  \valbest{\textbf{4.3$^+$}} &  \valbest{\textbf{4.9$^+$}} \\
\texttt{davinci-003} &           3.1$^-$ &        3.4\phantom{$^-$} &        3.9\phantom{$^-$} &        4.5\phantom{$^-$} \\
\texttt{GPT 3.5}          &  \valgood{\textbf{3.7$^+$}} &        \valgood{3.9\phantom{$^-$}} &  \valgood{\textbf{4.2$^+$}} &  \valgood{\textbf{4.9$^+$}} \\
\texttt{GPT 4}            &        3.7\phantom{$^-$} &        3.9\phantom{$^-$} &        4.1\phantom{$^-$} &        4.5\phantom{$^-$} \\
Human            &        3.5\phantom{$^-$} &        3.7\phantom{$^-$} &        3.9\phantom{$^-$} &        4.4\phantom{$^-$} \\\midrule
        \multicolumn{5}{c}{Social Bias Inference (SBIC)} \\ \midrule
    Model & Faithful & Relevant & Coherent & Fluent \\ \midrule 
Baseline         &           1.9$^-$ &           2.1$^-$ &           2.1$^-$ &           1.9$^-$ \\
\texttt{ada-001}     &        2.4\phantom{$^-$} &           2.2$^-$ &        2.7\phantom{$^-$} &  \textbf{3.3$^+$} \\
\texttt{babbage-001} &        3.1\phantom{$^-$} &        3.1\phantom{$^-$} &  \textbf{3.6$^+$} &  \textbf{3.8$^+$} \\
\texttt{curie-001}   &        3.4\phantom{$^-$} &        3.3\phantom{$^-$} &  \textbf{3.9$^+$} &  \valgood{\textbf{4.5$^+$}} \\
\texttt{davinci-001} &        3.4\phantom{$^-$} &        3.4\phantom{$^-$} &  \textbf{3.8$^+$} &  \textbf{3.9$^+$} \\
\texttt{davinci-002} &  \textbf{3.7$^+$} &        3.5\phantom{$^-$} &  \textbf{4.1$^+$} &  \textbf{4.2$^+$} \\
\texttt{davinci-003} &        3.5\phantom{$^-$} &        3.4\phantom{$^-$} &  \textbf{4.1$^+$} &  \textbf{4.4$^+$} \\
\texttt{GPT 3.5}          &  \valgood{\textbf{4.0$^+$}} &  \valgood{\textbf{3.7$^+$}} &  \valgood{\textbf{4.2$^+$}} &  \textbf{4.2$^+$} \\
\texttt{GPT 4}            &  \valbest{\textbf{4.1$^+$}} &  \valbest{\textbf{3.8$^+$}} & \valbest{\textbf{4.2$^+$}} &  \valbest{\textbf{4.6$^+$}} \\
Human            &        2.9\phantom{$^-$} &        3.0\phantom{$^-$} &        3.1\phantom{$^-$} &        2.6\phantom{$^-$} \\ \midrule
        \multicolumn{5}{c}{Annotator Backgrounds} \\ \midrule
    Task & \multicolumn{2}{c}{Education} & \multicolumn{2}{c}{Profession} \\ \midrule 
    \textsc{CovidET} & \multicolumn{2}{c}{MS,} & \multicolumn{2}{c}{CDC Health} \\ 
    & \multicolumn{2}{c}{Health Ed.} & \multicolumn{2}{c}{Comm. Specialist}\\ \hline
    MRF & \multicolumn{2}{c}{BA,} & \multicolumn{2}{c}{Grad Student,} \\
    & \multicolumn{2}{c}{Poli. Sci.} & \multicolumn{2}{c}{Public Policy}\\ \hline
    FLUTE & \multicolumn{2}{c}{MFA,} & \multicolumn{2}{c}{Writing Expert,} \\
    & \multicolumn{2}{c}{Creat. Writing} & \multicolumn{2}{c}{Grammarly}\\ \hline
    SBIC & \multicolumn{2}{c}{BS,} & \multicolumn{2}{c}{Grad Student,} \\
    & \multicolumn{2}{c}{Journalism} & \multicolumn{2}{c}{Epidemiology}\\ \hline
    Reframing & \multicolumn{2}{c}{BA,} & \multicolumn{2}{c}{Clinical Behavioral} \\
    & \multicolumn{2}{c}{Psychology} &\multicolumn{2}{c}{Health, Nurse}\\ \hline
    
    \end{tabular}
    }
    \caption{\revision{\textbf{Expert Scoring Evaluations for Zero-shot Generation Tasks} show that leading generative models (\texttt{davinci-003}, \texttt{GPT 3.5}) can match or exceed the faithfulness, relevance, coherence, and fluency of both fine-tuned models (Baseline) and gold references (Human). All scores are average ratings on 1-5 Likert scales.} Best models are in \valbest{\tiny green} followed by \valgood{\tiny blue}. Marks for $^+$ and $^-$ show performance significantly better or worse than human ($P < .05$) by Paired Bootstrap.}
    \label{tab:generation_finegrained_results}
\end{table}
\paragraph{Human Ranking Evaluation} According to the authors' ranking evaluations in Table~\ref{tab:generation_reliability_results},  \textbf{prompted \llm{s} produce helpful and informative text in all five generation tasks.} Model generations outrank the dataset's gold human reference at least 38\% of the time. The best models approach parity with humans---a near 50-50 coin toss to decide which is preferred. Furthermore, we see significant performance benefits from both RLHF models, \texttt{gpt-3.5-turbo} and \texttt{text-davinci-003}. Unlike classification (\S\ref{subsub:rq2}), our selected generation tasks seem to systematically benefit from human feedback.

Despite strong performances, no model substantially outperforms human annotation. This suggests that current \textbf{\llm{s} cannot fully replace human analysis}. Still, \revision{given \llm{'s} performance parity with humans, the results suggest one avenue for human-AI collaboration:} instead of coding text with summary explanations from scratch, researchers and annotators could apply minor edits to correct model generations. \footnote{Note that machine generated explanations might be limited in terms of their diversity. Although human validation can help refine these machine outputs, such process may not be able to introduce novel edits or perspectives.} The results in Table~\ref{tab:generation_reliability_results} suggest that, for every five model generations, 2 to 3 of these outputs would demand no additional annotator effort. \revision{If implemented successfully, this partnership could} significantly increase the efficiency of the social scientist's research pipeline. \revision{However, we leave it to future work in HCI to determine the plausibility of this partnership and the degree to which it might reduce annotators' cognitive load on exploratory analysis and free-coding tasks.}

As a tradeoff for \llm{'s} efficiency, \textbf{researchers will face the burden of manually validating generative outputs.} It is well-known that automatic performance metrics fail to capture  human preferences~\citep{goyal2022news,liang2022holistic}. In fact, we found that BLEU, BERTScore, and BLEURT, which rely on comparisons to human written ground truth, all produced uninterpretable scores for generation tasks. This highlights a fundamental challenge for evaluation of generation systems in CSS, especially if zero-shot performance continues to improve.
As zero-shot models approach or outperform the quality of the gold-reference generations, reference-based scoring becomes an \emph{invalid construct} for measuring models' true utility~\citep{raji2ai}, even if we assume the semantic similarity metrics are ideal.  
This motivates our use of reference-free expert evaluation of generations, that is, asking expert annotators which generation is more accurate with regard to the input. However, this alternative is limited by both cost and reproducibility concerns~\citep{karpinska-etal-2021-perils}. There is a clear need for new metrics and procedures to quantify model utility for CSS. 

\begin{table}[]
    \centering
    \resizebox{0.8\textwidth}{!}{%
    \begin{tabular}{l|rrrrr}
    \toprule
    & \multicolumn{5}{c}{\% Model Preferred Over Gold Annotations}\\
        Model & MRF & FLUTE & SBIC & Reframing & \textsc{CovidET} \\ \midrule 
        Baseline & 31.2\%  & 4.6\%  & 16.5\% & 45.0\%  & 37.5\%  \\ 
\texttt{ada-001} & 17.6\% & 1.7\% & 11.8\% & 0.0\%  & 23.5\% \\ 
\texttt{babbage-001} & 29.4\% & 6.7\% & 29.4\% & 0.0\% & 23.5\% \\ 
\texttt{curie-001} & 29.4\%  & 1.7\%  & 32.4\% & 11.5\% & 41.2\% \\ 
\texttt{davinci-001} & 21.4\% & 6.2\% & 39.0\% & 30.4\% & 50.0\% \\ 
\texttt{davinci-002} & 21.4\% & 25.0\% & 29.3\% & 10.0\% & 37.5\% \\ 
\texttt{davinci-003} & \valbest{38.9\%} & \valgood{47.0\%} & 50.0\% & \valgood{48.5\%} & \valgood{59.1\%} \\ 
\texttt{GPT 3.5} & 27.8\% & 37.9\% & \valbest{65.9\%} & \valbest{56.1\%} & \valbest{68.2\%} \\ 
\texttt{GPT 4} & \valgood{36.4\%} & \valbest{51.5\%} & \valgood{60.6\%} & 39.4\% & 36.4\% \\
    \end{tabular}
    }
    \caption{\textbf{Ranking Evaluations for Zero-shot Generation Tasks} give the proportion of all pairwise rankings where authors unanimously ranked the model's generation as more accurate or preferable to a gold-standard explanation drawn from the dataset. Best models are in \valbest{\tiny green} and runner-ups are in \valgood{\tiny blue}.
    }
    \label{tab:generation_reliability_results}
\end{table}

\section{Discussion}
\label{sec:discussion}
This work presents a comprehensive evaluation of \llm{s} on a representative suite of CSS tasks. We contribute a robust evaluation pipeline, which allows us to benchmark performance alongside supervised baselines on a wide range of tasks. Our research questions and empirical results are designed to help CSS researchers make decisions about when \llm{s} are suitable and which models are best suited for different research needs. In summary, we find that \textbf{\llm{s} can augment but not entirely replace the traditional CSS research pipeline.} 

More concretely, we make the following \textbf{recommendations to CSS researchers}:
\begin{enumerate}
    \item Integrate \llm{s}-in-the-loop to transform large-scale data labeling.
    \item Prioritize open-source \llm{s} for classification 
    \item \revision{Prioritize faithfulness, relevance, coherence, and fluency in your generations by opting for larger instruction-tuned models that have learned human preferences.}
    \item Investigate how \llm{s} produce new CSS paradigms built on the multipurpose capabilities of \llm{s} in the long term. 
\end{enumerate}

\begin{figure}
    \centering
    \includegraphics[width=\textwidth]{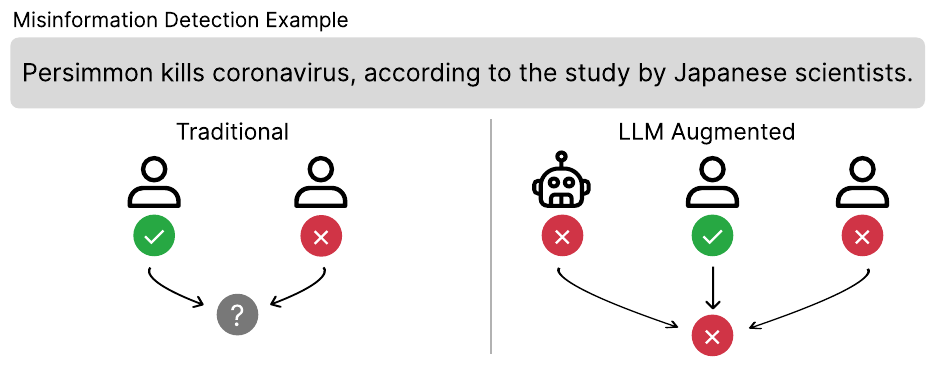}
    \caption{\textbf{Human-AI Collaboration} can improve the efficiency and reliability of text analysis. In this misinformation example, the \llm{} helps scale up annotation while reducing variance in the gold labels. Human annotation serves as validation for model-provided annotations.}
    \label{fig:mixed_annotation}
\end{figure}

In the following subsections, we more specifically detail in how annotation fits into the CSS pipeline (\S\ref{subsec:css_pipeline_regression}), how \llm{s} can augment annotation (\S\ref{subsec:css_pipeline_augment_annotation}), which \llm{s} we recommend for this purpose (\S\ref{subsec:css_pipeline_llm_choice}), and the opportunities \llm{s} expose for new research paradigms (\S\ref{subsec:css_pipeline_blended_paradigm} and \S\ref{subsec:new_eval_paradigm}).

\subsection{How Annotation Fits Into CSS}
\label{subsec:css_pipeline_regression}
\revision{Social scientists are not often interested in classification labels or generative codes merely for their own sake. Labeled text is almost always used to explain a wider phenomenon using downstream inferential statistics such as regression. To ensure a valid downstream inference, any estimators of the underlying construct need to be asymptotically unbiased and allow for the computation of valid confidence intervals. In a study that follows up our own, \citet{egami2023using} demonstrate that even the most accurate of \llm{} annotations will produce biased estimates and invalid confidence intervals when they are merely averaged. However, \citet{egami2023using} propose a method called Design-based Semi-supervised Learning (DSL) which they use for computing unbiased estimators for prevalence estimation on each of our 17 classification tasks. Using FLAN-UL2 pseudo-labels and only a handful of gold-labeled instances, they find that it is possible to compute unbiased regressions, even for tasks with low accuracy from UL2. With only 25 gold annotations, DSL can compute confidence intervals that have $>$86\% correct coverage for all of our classification tasks \textit{including those with the lowest performance from \llm{} pseudo-labels.}}

\subsection{\llm{s} Can Augment Annotation}
\label{subsec:css_pipeline_augment_annotation}
\textit{Our work shows that current \llm{s} perform well enough to augment the CSS annotation pipeline and thus reduce the need for human labor on some tasks.} \S\ref{subsub:rq1} show that an \llm{} annotator yields fair to strong agreement with humans on 12 out of 17 classification tasks.
However, \textit{\llm{s} are not a wholesale replacement for human annotators}. Even the best \llm{s} exhibit unusably low performance on some CSS tasks. Ensembling prediction does not mitigate this label corruption as \llm{s} demonstrate high internal agreement, even when inaccurate~\citep{gilardi2023chatgpt}. Overconfident models, if left unchecked, distort the conclusions of CSS research and subsequently mislead policy and social actions taken in response. Human validation is key to avoiding a replication crisis in CSS caused by \llm{} hallucinations and inaccuracies.

Instead, \textit{we advocate that CSS researchers integrate \llm{s} with human annotation}, as illustrated in Figure~\ref{fig:mixed_annotation}. When a \llm{} matches human levels of agreement, it can be used as one of multiple labelers. It is possible from such methods to produce unbiased estimators. The estimation methods of \citet{chaganty-etal-2018-price} are provably optimal, and provide these unbiased estimators with 7-13\% monetary savings by leveraging \llm{} pseudo-labels.

Moving forward, \llm{s} can serve as a \textit{flywheel} for dataset collection. Prompted \llm{s} consistently perform significantly better than chance, providing imperfect labels at low cost. Annotation schemes developed to iteratively improve imperfect data ---such as weak supervision~\citep{ratner2017snorkel}, targeted data cleaning~\citep{chen2022clean}, and active learning~\citep{yuan-etal-2020-cold, li2022pre} --- avoid \llm{} pitfalls by allowing human validation to refine the original model.  This creates a virtuous cycle which exploits the strengths of \llm{s} to focus human expertise where it is most needed~\citep{kiela-etal-2021-dynabench}. 

Our results show that \textit{\llm{s} are even \revision{stronger at} generation tasks}, being rated superior to human gold annotations over 38\% of the time in all 5 tasks we evaluate. \llm{s} can already generate syntactically cohesive and stylistically consistent text, and as we have shown, the best generations are also highly relevant and faithful to core CSS tasks. Humans expertise can be used to further curate outputs according to domain-specific accuracy and quality metrics. Dataset construction through human curation of \llm{} generations has already emerged in recent NLP works on decision explanation~\citep{wiegreffe-etal-2022-reframing}, model error identification~\citep{ribeiro2022adaptive}, and even to build the figurative language benchmark used in this work~\citep{chakrabarty2022flute}.

We recommend that CSS researchers consider the use of \llm{s} as the foundation of such annotation procedures, \revision{and that future studies measure the degree to which this strategy improves annotation efficiency as we suspect. If the anticipated efficiency is achieved,} CSS researchers should reinvest savings to train \textbf{expert annotators}, reversing the trends of replacing experts with crowdworkers due to cost~\citep{snow-etal-2008-cheap}. By doing so, \llm{s} could enable data labeling procedures that more deeply benefit from the non-computational expertise of the social scientists whose theories we build upon.

\subsection{When To Use What \llm{}}
\label{subsec:css_pipeline_llm_choice}

We hope these results help CSS researchers to understand \llm{} alternatives for their use cases. Our general prompt guidelines allow us to quickly design functional prompts for many models. When looking to incorporate \llm{s} in their work, CSS researchers should consider the advantages and disadvantages of open and closed-source models. 

\textit{For CSS classification, our work shows that open-source models like FLAN are as capable as state-of-the-art \revision{closed-source} \llm{s} from OpenAI.} We recommend researchers who already have access to GPUs capable of running these models prefer \revision{open-source} models. For continuous monitoring and enormous-scale analysis, the low marginal cost of these models could make them price-advantageous. For CSS researchers with expertise, open-source \llm{s} have the added benefit of being able to be fine-tuned on labeled data and constrained programatically for more predictable behavior. At this time, it is not possible to further fine-tune all of OpenAI's instruction-tuned models. 

For those without existing hardware infrastructure, \revision{proprietary APIs appear to be} a cost-efficient option. Based on current cloud pricing,\footnote{\href{https://medium.com/google-cloud/deploy-flan-t5-xxl-on-vertex-ai-prediction-579953afdc88}{Google Cloud FLAN hosting cost}} the hardware necessary to run FLAN-T5-XXL costs 170 dollars per day---the equivalent of processing $\sim$50 million words with \texttt{gpt-3.5-turbo}.\footnote{\href{https://openai.com/pricing}{OpenAI Pricing}} In most cases, \texttt{gpt-3.5-turbo} is more cost-efficient and has a lower operational overhead for hardware-constrained research groups.

For generation tasks, the results are clear-cut. Even the largest open-source models failed to generate meaningful responses for CSS tasks. Even when labeled data is available, the best \textit{\revision{proprietary} models outperform fine-tuned baselines consistently} and approach parity with gold human annotations when evaluated by crowdworkers. For CSS experts looking to generate interpretations or explanations of data, \texttt{gpt-3.5-turbo} is the clear leading \llm{} by both price and performance. No matter which modeling decision is made, practitioners should keep the limitations of natural language generation in mind, understanding that explanations are not causal and recognizing the risks that come with model errors and hallucinations (see \S\ref{subsub:final_considerations_for_css}). 

Our work shows that \textbf{all \llm{s} struggle most with conversational and full document data}. Also, \textbf{\llm{s} currently lack clear cross-document reasoning capabilities}, limiting extremely common CSS applications like topic modeling. For CSS subfields which often study these discourse types---sociology, literature, and psychology---\llm{s} have major limitations and are unlikely to have major immediate impact. NLP researchers who aim to improve existing \llm{s} to empower more CSS tasks should study the unique technical challenges of conversations, long documents, and cross-document reasoning~\citep{beltagy2020longformer, caciularu2021cdlm, yu2021score}.

\subsection{Blending CSS Paradigms}
\label{subsec:css_pipeline_blended_paradigm}
The few-shot~\citep{brown2020language} and zero-shot capabilities~\citep{ouyang2022training} of \llm{s} \textbf{blur the traditional line between supervised and unsupervised ML methods for the social sciences}. Historically, supervised methods invest in labeled data guided by existing theory to develop a trained model. This model is then used to classify text at scale to gather evidence for the causal effects surrounding the theory. By comparison, unsupervised methods like topic modeling often condense large amounts of information to help researchers discover new relationships, which develop or refine social theories \citep{evans2016machine}. 

The ability of \llm{s} to follow instructions and interpret complex tasks is rapidly advancing, with major new models even within the course of this work~\citep{gpt4}. Beyond annotation, \llm{s} have multi-purpose capabilities to retrieve, label, and condense relevant information at scale. We believe that this can blend the boundaries between supervised and unsupervised paradigms. Rather than using separate paradigms to develop and test theories, a single tool can be used to develop working hypotheses, using generated and summarized data, and test hypotheses, labeling human samples flexibly with low-cost classification capabilities. We believe CSS researchers should use the multi-functionality of \llm{s} to create new paradigms of research for their fields.

\paragraph{Simulation} An emerging example of such innovation in CSS is the use of \llm{s} as simulated sample populations. Game theorists have used rule-based utility functions to develop hypotheses about the causes of social phenomena~\citep{schelling1971dynamic, easley2010networks} and to predict the effects of policy changes~\citep{shubik1982game, kleinberg2018human}. However, simulations are limited by the expressiveness of utility functions~\citep{ellsberg1961risk, machina1987choice}. LLMs hold a great potential to provide more powerful simulations for CSS \citep{bail2023can}, as they replicate human biases without explicit conditioning~\citep{jones2022capturing, koralus2023humans}. Recent work leverages this capacity to simulate social computing systems~\citep{park2022social}, community and their members' interactions \citep{park2023generative}, public opinion~\citep{argyle2022out, chu2023language}, and subjective experience description~\citep{argyle2022out}. 

However, there are \textit{dangers and uncertainties} in this area as noted in these works. Since social systems evolve unpredictably~\citep{salganik2006experimental}, simulated samples inherently have limited predictive and explanatory power. While utility-based simulations have similar limitations, their assumptions are explicit unlike the opaque model of human behavior an \llm{} provides. Additionally, current models exhibit higher homogeneity of opinions than humans~\citep{argyle2022out, santurkar2023opinions}. Combining \llm{s} with true human samples is essential to avoid an algorithmic \emph{monoculture} and could lead to fragile findings covering only the limited perspectives represented~\citep{kleinberg2021algorithmic, bommasani2022picking}. 

\subsection{The Need for A New Evaluation Paradigm} 
\label{subsec:new_eval_paradigm}
Evaluation will need to adapt if blended methods create a new CSS paradigm. Accuracy-based metrics were ideal for fixed-taxonomy classification tasks in the era of NLP benchmarking. Similarly, word-overlap metrics made sense for natural language generation tasks in which the gold reference was well-defined (e.g., translation). However, open-ended coding and CSS explanation objectives follow neither a pre-defined taxonomy nor a regular output template. For more open-ended data exploration tasks like topic modeling, held-out likelihood helped automatically measure the predictive power of the model \citep{wallach2009evaluation}, but predictiveness does not always correlate with explainability \citep{shmueli2010explain}, and these automatic metrics proved to be at odds with human quality evaluations \citep{chang2009reading}. In CSS, human evaluations can be unreliable \citep{karpinska-etal-2021-perils}. We observe this directly in our work, as crowd work seems to provide unreliable quality metrics for FLUTE, a nuanced generative task. New metrics are needed to capture the semantic validity of free-form coding with \llm{s} as explanation-generators. 

\subsection{CSS Challenges for \llm{s}}
\label{subsec:css_challenges}
As shown by our \S\ref{subsec:clf_results} results, \llm{s} face notable challenges that pervade the computational social sciences. The first challenge comes from the subtle and non-conventional language of \textbf{expert taxonomies}. Expert taxonomies contain technical terms like the dialect feature \textit{copula omission} (\S\ref{subsub:dialect_feature_detection}), plus specialized or nonstandard definitions of colloquial terms, like the persuasive \textit{scarcity} strategy (\S\ref{subsub:utterance_persuasion}), or \textit{white grievance} in implicit hate (\S\ref{subsub:utterances_hate_speech}). \llm{s} may lack sufficient representations for such technical terms, as they may be absent from the pretraining data \citep{yao2021adapt}.  How to \emph{teach} \llm{s} to understand these social constructs deserves further technical attention.  This is especially true for \textit{novel theoretical constructs} that social scientists may wish to define and study in collaboration with \llm{s}.

Unlike widely used NLP classification tasks, the challenge of expert taxonomies in CSS is compounded by the \textbf{size of the target label space}, which, in CSS applications, may contain upwards of 72 classes (see \textit{character tropes}, \S\ref{subsub:role_tagging}). This challenges transformer-based \llm{s,} which have relatively limited memory, finite processing windows, and quadratic space complexity.

Large, complex, and nuanced annotation schemes may also introduce dependencies among labels that are organized into multi-level hierarchies or richly constrained schemas, as in many \textit{event argument extraction} applications.
Such complex \textbf{structural parsing} tasks pose special challenges to the zero-shot prompting paradigm introduced in this work since prompted models often struggle to generate \textbf{consistent outputs} \citep{mishra2019modular}. Our prompting best practices in Table~\ref{tab:prompting_guidelines} all help \llm{s} generate more consistent machine-readable outputs, but this challenge is not fully solved for all CSS tasks.

Finally, Computational Social Scientists study language, norms, beliefs, and political structures that all \textit{change across time}. To account for these distribution shifts, \llm{s} will need an extremely high level of \textbf{temporal grounding}---knowledge and signals by which to orient a text analysis in a particular place and time \citep{bommasani2021opportunities}. This is especially challenging wherever researchers are interested in \textbf{rapid, synchronous analysis of breaking events}. It may be prohibitively expensive to frequently update \llm{'s} knowledge of current events via continually training \citep{bender2021dangers}, and this challenge will only be exacerbated as models continue to scale up. 

\subsection{Issues in Bias and Fairness} 
\label{subsub:bias_fairness}
\paragraph{Bias} Researchers should weigh the benefits of applying prompting methods to CSS, along with the limitations and risks of doing so. Most notably, \llm{s} are known to amplify social biases and stereotypes \citep{sheng-etal-2021-societal,abid2021persistent,borchers-etal-2022-looking,lucy-bamman-2021-gender, shaikh2022second}, as well as viewpoint biases in subjective domains \citep{santurkar2023whose}. These biases can emerge in open-ended generation tasks like the explanation and paraphrasing \citep{dhamala2021bold}. The performance of \llm{s} as tools for classification and parsing may vary systematically as a function of demographic variation in the target population \citep{zhao2018gender}. With the datasets available, we were unable to perform a systematic analysis of biases and performance discrepancies, but we urge researchers to carefully consider these risks in downstream applications.

\paragraph{Risks Inherent to Proprietary \llm{s}} \revision{Researchers will often lack details on the selection, filtering, and formatting of online corpora for training proprietary models like OpenAI's GPT-4. There are unique risks and privacy implications for those who employ these models for research. One risk inherent in an obscured pre-training corpus is that we can't decompose the above issues of bias and social harm into their respective sources for targeted mitigation. Especially with open-ended generative coding (\S\ref{subsec:generation}), unattributed biases could jeopardize the reliability and prosocial impact of downstream scientific analyses.} Such issues may also escape human review, as humans are prone to falsely attribute \textit{factuality} to \revision{texts that bear a more authoritative or expert \textit{style} \citep{wu2023style}----a style that modern proprietary \llm{s} have largely mastered. Furthermore, black-box industrial APIs may not allow for targeted mitigation via fine-tuning.}

\revision{Researchers who operate with closed-source industrial APIs may also be more prone to privacy issues and legal disputes over intellectual property. Proprietary models are known to replicate copyrighted materials and sensitive personally identifiable information \citep{carlini2021extracting}. Researchers who employ these models may be unknowingly accountable for knowledge based on false or missing attributions as well as private personal information.}

\subsection{Limitations}
\label{sec:limitations}

\paragraph{Task Selection and Data Leakage}
\label{subsub:ethical_considerations}
Our tasks do not represent an exhaustive list of all application domains. Some highly-sensitive domains like mental health \citep{nguyen2022improving}, which requires expert annotations, and cultural studies, which requires community-specific knowledge, are rife with additional challenges and ethical concerns. These are largely outside the scope of the current study. More broadly, \llm{s} should not be used to give legal or medical advice, prescribe or diagnose illness, or interfere with democratic processes \citep{solaiman2021process}. Finally, our task selection was limited by the available data resources in the field, which is largely dominated by text in standard dialects \citep{ziems-etal-2023-multi} representing members of Western, Educated, Industrial, Rich, and Democratic populations \citep[WEIRD;][]{ignatow2016text, muthukrishna2020beyond}. Future studies should separately consider LLMs' utility for cross-cultural CSS.

When evaluating \llm{s}, one notable concern is data leakage. Data from the test set might have been seen by \llm{s} during the pre-training, and this would artificially inflate test performances. This problem is especially concerning for closed-source or proprietary models with undisclosed training sets. One mitigation strategy is to design explicit prompts that force the model to forget the test set. Another strategy is to design custom test sets from perturbations of existing data to more fairly evaluate models. We leave this for future work.

\paragraph{Causality and Explanations}
\label{subsub:final_considerations_for_css}
Explanations are important to social science \citep{shmueli2010explain,hofman2017prediction,yarkoni2017choosing}. In this work, we explored the predictive power of \llm{s} rather than causal explanations. Predictions serve to expose and elaborate on the underlying social phenomena latent in a text. These explicit phenomena can then be used as structured features for further causal analysis.

However, this may not be sufficient: social scientists often seek causal theories \citep{dimaggio2015adapting}, or at least \textit{contrastive} explanations, \textit{why $P$ instead of $Q$} \citep{miller2019explanation}. Because \llm{s} are not grounded in a causal model of the world \citep{bender2021dangers}, they are not on their own reliable tools for mining causal relationships in text. We leave it to future work to explore contrastive or causal explanations in \llm{s}.

\begin{acknowledgments}
We are thankful to Tony Wang, Rishi Bommasani, Albert Lu, Myra Cheng, Yanzhe Zhang, Camille Harris, and Minzhi Li for their helpful feedback on our early drafts. We thank the anonymous reviewers for their insightful suggestions as they significantly shaped the final form of this work. Caleb Ziems is supported by the NSF Graduate Research Fellowship under Grant No. DGE-2039655. This work was partially sponsored by NSF grant IIS-2247357 and IIS-2308994.

\end{acknowledgments}

\starttwocolumn
\bibliography{custom}

\end{document}